\documentclass[a4paper,fleqn]{cas-dc}

\usepackage[authoryear]{natbib}
\usepackage{amsmath}
\usepackage{subfigure}
\usepackage{booktabs}
\usepackage{multirow}
\usepackage{graphicx}
\usepackage{float}
\usepackage{url}
\usepackage{adjustbox}
\usepackage[ruled, vlined, linesnumbered]{algorithm2e}
\usepackage{caption}
\usepackage{subcaption}
\usepackage{hyperref}

\newcommand{\eqrangeref}[2]{Eq.~\eqref{#1}-\eqref{#2}}

\def\tsc#1{\csdef{#1}{\textsc{\lowercase{#1}}\xspace}}
\tsc{WGM}
\tsc{QE}
\usepackage{ulem} 
\usepackage[switch]{lineno}
\begin{document}

\let\WriteBookmarks\relax
\def\floatpagepagefraction{1}
\def\textpagefraction{.001}
\let\printorcid\relax



\title [mode = title]{Segmentation-aware Prior Assisted Joint Global Information Aggregated 3D Building Reconstruction}  



%

\author[1, 4]{Hongxin Peng}[style=chinese]

\author[2]{Yongjian Liao}[style=chinese]
\fnmark[1]

\author[3]{Weijun Li}[style=chinese]

\author[1]{Chuanyu Fu}[style=chinese]

\author[1]{Guoxin Zhang}[style=chinese]

\author[1]{Ziquan Ding}[style=chinese]

\author[1]{Zijie Huang}[style=chinese]

\author[3]{Qiku Cao}[style=chinese]

\author[4]{Shuting Cai}[style=chinese]
\cormark[1]
\ead{shutingcai@gdut.edu.cn}

\cortext[1]{Corresponding authors.}
\fntext[1]{Equivalent to the first author’s contribution to this work.}

\address[1]{School of Automation, Guangdong University of Technology, Guangzhou 510006, China}
\address[2]{School of Artifcial Intelligence and Automation, Huazhong University of Science and Technology, Wuhan 430074, China }
\address[3]{School of Computer Science, Guangdong University of Technology, Guangzhou 510006, China}
\address[4]{School of Integrated Circuits, Guangdong University of Technology, Guangzhou 510006, China}















\begin{abstract}
Multi-View Stereo plays a pivotal role in civil engineering by facilitating 3D modeling, precise engineering surveying, quantitative analysis, as well as monitoring and maintenance. It serves as a valuable tool, offering high-precision and real-time spatial information crucial for various engineering projects. However, Multi-View Stereo algorithms encounter challenges in reconstructing weakly-textured regions within large-scale building scenes. In these areas, the stereo matching of pixels often fails, leading to inaccurate depth estimations. Based on the Segment Anything Model and RANSAC algorithm, we propose an algorithm that accurately segments weakly-textured regions and constructs their plane priors. These plane priors, combined with triangulation priors, form a reliable prior candidate set. Additionally, we introduce a novel global information aggregation cost function. This function selects optimal plane prior information based on global information in the prior candidate set, constrained by geometric consistency during the depth estimation update process. Experimental results on both the ETH3D benchmark dataset, aerial dataset, building dataset and real scenarios substantiate the superior performance of our method in producing 3D building models compared to other state-of-the-art methods. In summary, our work aims to enhance the completeness and density of 3D building reconstruction, carrying implications for broader applications in urban planning and virtual reality.
\end{abstract}



\begin{keywords}
 Multi-View Stereo \sep Large-Scale building scenes \sep Weakly-Textured regions  \sep Segment Anything Model \sep 3D reconstruction \sep Global information aggregation 
 
\end{keywords}

\maketitle

\section{Introduction}
3D reconstruction is a popular research area in computer vision, particularly in engineering application, focusing on reconstructing precise 3D building models of architectural scenes. 3D reconstruction has played a critical role in engineering fields such as civil engineering, architecture, construction, and facility management \citep{r43,r41}. Multi-View Stereo (MVS) techniques are pivotal in achieving detailed 3D reconstruction from multiple images, especially in complex architectural environments. MVS algorithms utilize images captured from different perspectives to accurately reconstruct 3D models by leveraging robust matching algorithms and photometric consistency principles. This approach not only enhances the fidelity and completeness of 3D models but also supports applications in construction planning \citep{r55} and progress controlling \citep{r56}, addressing challenges such as cost reduction, efficiency improvement, and safety enhancement in modern architectural practices. However, due to the high accuracy and completeness requirements for the reconstructed 3D models, MVS still faces challenges, including issues such as weakly-textured regions, occluded objects, and changes in lighting. These challenges result in insufficient completeness of the 3D point cloud.
\par
Currently, as a key technology in 3D reconstruction, most MVS algorithms rely on photometric consistency for pixel matching across multiple views. These algorithms achieve higher matching accuracy and finer reconstruction quality in regions with strong textures due to their richness and stability in feature points within these areas. However, in weakly-textured regions, such as walls, floors, roofs, and building surfaces, the pixel information within the matching window becomes highly homogeneous, posing challenges in pixel matching. We term this situation the matching ambiguity problem. Addressing this issue is intricate, as reliance on plane hypotheses of neighboring pixels may lead to being trapped in local optima within the propagation mechanism. Consequently, this results in a decrease in both accuracy and completeness of the 3D model, particularly noticeable when reconstructing with high-resolution images.
Additionally, the light spots and shadows generated by multi angle light sources further exacerbate the matching ambiguity problem, and even cause the 3D model missing point clouds in weakly-textured regions. This issue is a significant challenge in both engineering applications and academic research. 
\begin{figure*}[t]
    \centering
    \includegraphics[width=\linewidth]{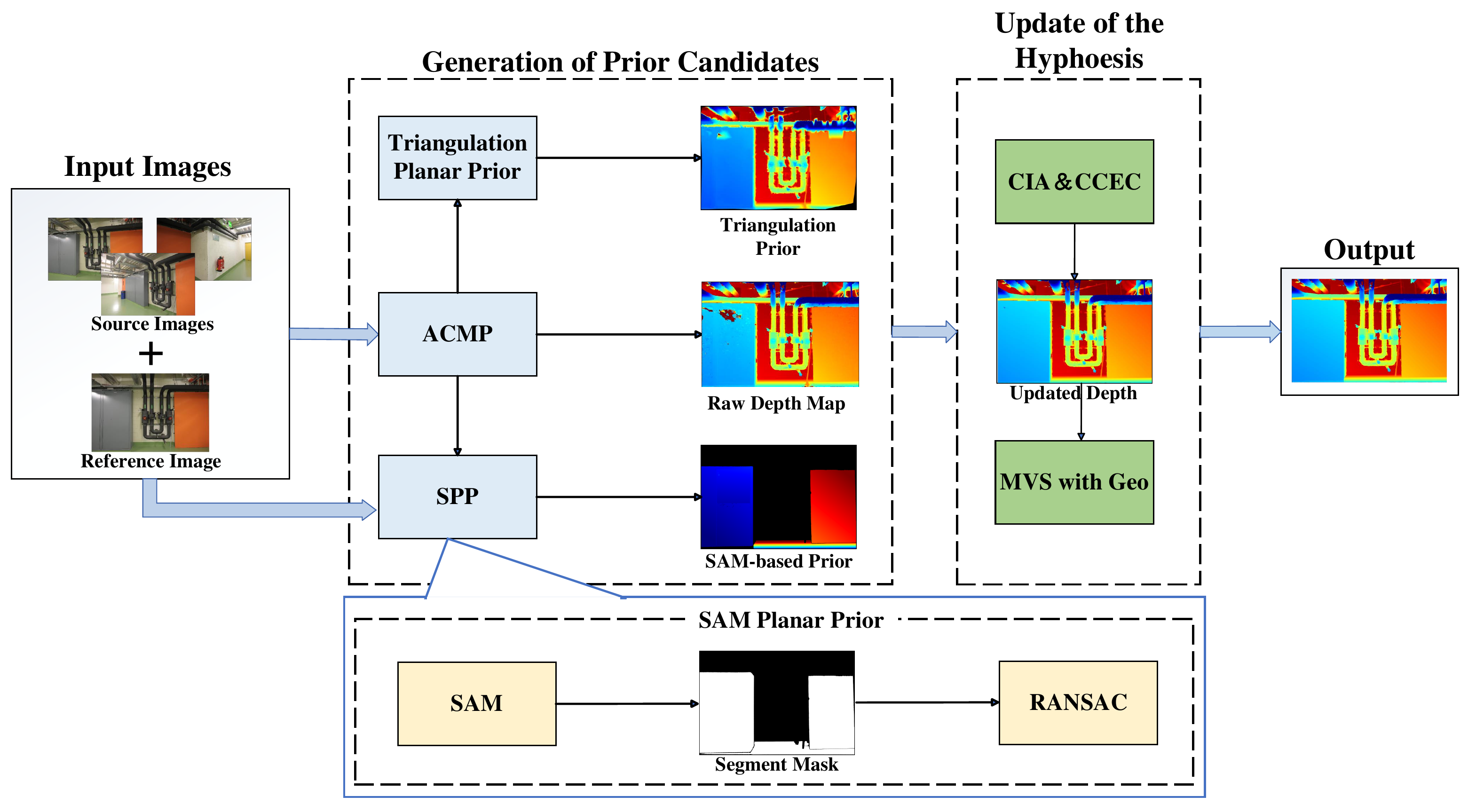}
    \caption{Overall workflow of the proposed approach. In the first stage, we generate raw depth from ACMP algorithm and the plane prior candidates composed of Delaunay triangulation prior and SAM Planar Prior (SPP). Then, we embedding prior candidates set into the update of depth estimation employing global information aggregation cost function (GIA), along with Geometric consistency for epipolar constraints (CGEC). Finally, we obtain the final depth estimation after PatchMatch MVS with geometric consistency.}
    \label{fig3}
\end{figure*}
\par
Some solutions based on the PatchMatch algorithm \citep{r15} have utilized planar information in the scene to calculate plane priors, aiming to address the limitations of photometric consistency. \cite{r8} propose a PatchMatch algorithm assisted by plane priors, preserving low-cost plane hypotheses and generating prior planes through the Delaunay triangulation algorithm for plane segmentation. However, due to the presence of local optima in photometric consistency, there are cases where points have low cost values even though their depth estimation is incorrect. TAPA-MVS \citep{r29} assumes that all patches in weakly-textured regions are similar. It introduces candidate superpixel planes to the optimization framework, contributing in the depth estimation process. However, these methods excessively fragment the plane segmentation, neglecting global and semantic information about the planes. This has resulted in many prior planes still being incorrect. Consequently, the reconstructed planes lack sufficient smoothness and may still contain holes. Especially in weakly-textured regions of buildings, there is a situation where inaccurate depth estimation leads to point cloud missing.
\par
To address the issues in the aforementioned plane-prior-based MVS methods, this paper proposes an innovative PatchMatch algorithm based on prior information. The workflow of our method is illustrated in \autoref{fig3}. By integrating prior information from weakly-textured regions, we address problems such as fragmentation, lack of smoothness, and holes in plane priors. To obtain reliable prior information, we employ the Segment Anything Model \citep{r25} to semantically segment the image, yielding precise large plane segmentation masks for weakly-textured regions. The RANSAC \citep{r27} method and Delaunay triangulation method are used to fit different prior planes, forming a candidate set of prior planes. 
\par
In addition, in order to seamlessly integrate prior information into the depth estimation update process, we introduce a novel cost aggregation function based on global information. This function considers both photometric and geometric consistency while incorporating global information about planes. The proposed approach significantly mitigates the impact of unreliable prior information and noise, enhancing the robustness and accuracy of the depth estimation process. It also addresses the limitations posed by large and complex scenes in engineering applications, as well as the issue of information loss during scene reconstruction.
\par
The main contributions of our paper are summarized as follows:
\par
$\bullet$
We propose a prior plane generation algorithm based on the Segment Anything Model (SAM). This algorithm inherits SAM's strong generalization performance, accurately segmenting weakly-textured regions in different scenes and generating a comprehensive set of prior planes. This addresses issues such as excessive segmentation of large planes and insufficient surface smoothness.
\par
$\bullet$
We introduce a novel global information aggregation cost, employing a probabilistic graph model, to address matching ambiguity in weakly-textured regions. By optimizing geometric consistency, this cost reduces the impact of unreliable plane hypotheses or noise, facilitating a more effective integration of prior hypotheses into the depth estimation update process.
\par
$\bullet$
We utilize the SAM and Delaunay triangulation to form a candidate set of prior planes building upon PatchMatch-MVS. By computing the global information aggregation cost, we ensure the accuracy and completeness of depth estimation in weakly-textured regions within the scene.
\section{Related Work}

MVS is one of the most effective large-scale building scene reconstruction algorithms that aims to obtain the 3D depth information of a scene by performing stereo matching across multiple images. \cite{r48} propose a new MVS depth completion network, which reconstructs complete point clouds of buildings and civil infrastructure by completing missing areas in depth maps. This method has practical advantages in the field of 3D civil structure reconstruction and promotes the development of digital models of buildings and infrastructure in the Architecture, Engineering, and Construction (AEC) sector. \cite{r49} propose a new photogrammetric based Computational fluid dynamics (CFD) framework that integrates MVS-based 3D point cloud reconstruction to reconstruct structures from 2D images obtained from portable devices such as mobile phones and drones. And an immersogeometric approach
that can perform flow analysis directly on the reconstructed point cloud. \cite{r50} improve the PatchMatchNet algorithm \citep{r52} and Iterative Closest Point (ICP) algorithm \citep{r53} to realize intelligent measurement of full-field displacement of retaining wall structure and provide data support for working condition assessment of supporting structure and geological disaster warning.

\par
Traditional MVS matching algorithms can be categorized into four main types\citep{r35}: voxel-based methods \citep{r11,r36,r37}, grid-based methods \citep{r12}, feature point-based methods \citep{r13}, and depth map fusion-based methods \citep{r14,r38,r39,r40}. This paper focuses on introducing depth map fusion-based methods.
\par
Currently, depth map fusion for MVS matching algorithms is primarily based on the core idea of the PatchMatch algorithm. The primary objective of the PatchMatch algorithm is to identify corresponding pixel patches between two correlated images. Subsequently, several extended algorithms based on the PatchMatch idea have been proposed, including PatchMatch Stereo \citep{r16}, OpenMVS \citep{r17}, and ACMM \citep{r5}.
\par
For example, \cite{r18} introduce a pixel view selection probability model, alternatingly updating depth estimates and view selection probabilities through generalized expectation maximization. \cite{r19} introduce geometric consistency into the view selection process, selecting appropriate neighboring views and considering geometric consistency constraints in the matching cost calculation, significantly improving the accuracy of 3D reconstruction. However, these methods have long-term limitations for large and complex scenes, and there is also the problem of information loss in reconstructing the representation of shapes. Furthermore, due to the high accuracy and completeness requirements for the reconstructed 3D models, MVS still faces challenges in weakly-textured regions.
\par
To address the challenge of estimating depth in weakly-textured regions, some methods leverage the strong planarity assumption in these regions, which is based on the prior concept that weakly-textured regions are usually on the same plane. In the ACMP algorithm \citep{r8}, only depth values with high confidence is employed for planar fitting to generate a prior planar model, guiding the depth estimation. However, this method tends to overly segment planes and only considers local information, leading to the problem of planar fragmentation. Subsequently, ACMMP, as introduced by \citep{r20}, incorporates a pyramid structure based on the ACMP algorithm. It aims to diminish the number of segmented planes through downsampling, yet it does not fundamentally address the challenge of planar fragmentation. \cite{r21} utilize semantic information to identify weakly-textured planes in the scene and generated prior planes. Then, they incorporated this prior information into the matching cost calculation to guide depth estimation in weakly-textured regions.
TAPA \citep{r29} introduces candidate superpixel planes for joint PatchMatch. \cite{r54} extended this framework by making use of a outlier filtering in the disparity domain that used in SGM for superpixel clustering, this method can effectively reduce the impact of noise on plane fitting.
However, the segmentation accuracy of their segmentation model was insufficient, severely affecting the fitting of prior planes. To overcome these challenges, we propose a MVS algorithm that incorporates multiple prior planes and global constraints.
\begin{figure*}[t]
    \centering
    \includegraphics[width=\linewidth]{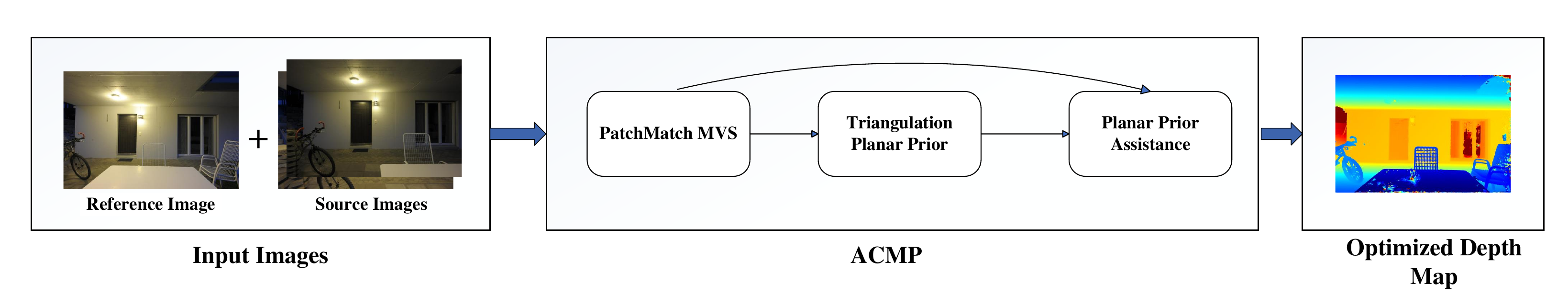}
    \caption{Overall workflow of ACMP. Firstly, PatchMatch MVS model is employed for the input images to generate raw depth. After sparsification, Delaunay triangulation is used to model planes and generate a triangulated prior plane. Subsequently, with the help of planar models, planar prior assistance is leveraged to optimize depth maps.}
    \label{figacmp}
\end{figure*}
\section{Preliminaries}
In this section, we review a PatchMatch MVS framework, ACMP \citep{r8}, which serves as the baseline for our method. The process can be divided into two parts. The workflow of ACMP is illustrated in \autoref{figacmp}. Firstly, PatchMatch-based MVS model is employed for the input images to generate raw depth. After sparsification, Delaunay triangulation is used to model planes and generate a triangulated prior plane. Subsequently, leveraging multi-view aggregation matching costs, selects the optimal hypothesis for each pixel, ultimately yielding the optimized depth estimation.

\subsection{PatchMatch MVS}
PatchMatch MVS consists mainly of four steps:
random initialization, propagation, multi-view matching 
cost evaluation, and refinement. The purpose of this process is to recover the depth map of the current image, i.e., the reference image, under the guidance of the source images, which are the neighboring images corresponding to the reference image. Patch Match MVS uses these source images to calculate multi-view costs to
optimize the depth hypotheses for each pixel in the reference image. In random initialization stage, 
the initial hypothesis with multi-view aggregated matching 
cost is randomly generated for each pixel in the reference 
image. Propagation involves sampling and propagating candidate hypotheses to neighboring pixels, spreading good 
hypotheses to the current pixel. Multi-view matching costs, which are computed from each of 
the source images via homograhy, 
assesses the sampled candidate hypotheses, assisting in the update of hypotheses. Refinement is employed to enrich the diversity of the solution space and select the 
optimal hypothesis estimate.
\subsection{Triangulation Planar Prior}
In the first stage, ACMP filters out some unreliable points in the raw depth maps through multi-view aggregation of photometric consistency costs, retaining relatively trustworthy points to form a sparse plane model. Then, the Delaunay triangulation is performed on these sparse points, connecting them into triangles, as illustrated in \autoref{fig7}(c). The plane information of the three vertices of each triangle determines the depth and normal vector estimates of the pixels within the triangle area, allowing the recovery of depth for the filtered-out points. For all pixels in the triangular region, their normal vector $ n_t=(A_{t},B_{t},C_{t})$ and the distance from the origin to triangulated plane $d_{t}$ can be calculated using the 3D coordinates of the triangle's three vertices according to the spatial plane formula: 
\begin{eqnarray}
A_{t}x_{i}+B_{t}y_{i}+C_{t}z_{i}+d_{t}=0,
\end{eqnarray}
where $(x_{i},y_{i},z_{i})$ represents the 3D coordinates of the triangle vertices. After obtaining the plane normal vector, we calculate the depth $D_{t}$ of the pixel $p_{t}(x_{t},y_{t})$ within the plane using the following formula: 
\begin{eqnarray}
D_{t}=-\frac{d_{t}}{A_{t}\frac{x_{t}-c_x}{f_x} +B_{t}\frac{y_{t}-c_y}{f_y}+C_{t}},
\end{eqnarray}
where $f_x$ and $f_y$ are the camera's focal lengths, and $c_x$ and $c_y$ are the optical center coordinates of the image. These parameters are all intrinsic parameters of the camera.

\subsection{Planar Prior Assistance}
In weakly textured regions, photometric consistency alone may not provide reliable depth estimates, while prior plane information offers additional cues. In regions with stronger textures, photometric consistency costs are reliable, but prior plane information might introduce errors. Therefore, the ACMP algorithm comprehensively considers both aspects to obtain a multi-view aggregation matching cost. This approach ensures the effective utilization of prior information in weakly-textured regions while maintaining the reliability of photometric consistency in regions with stronger textures. The multi-view aggregation matching cost is represented as follows:
\begin{eqnarray}
\label{eq33}
Cost_{agg}=\frac{Cost_{m-ph}^{2}}{\alpha}-\log[\gamma+e^{-\frac{(d_i-d_p)}{2\lambda_{d}}}e^{-\frac{\arccos^{2}n_{i}^{T}n_{p}}{2\lambda_{n}}}],
\end{eqnarray}
where $\alpha$ is the constant weight of the photometric consistency cost, and $\gamma$ is the penalty term for prior information. The first term of \autoref{eq33} is composed of photometric consistency, implying that in regions with stronger textures, the variations in photometric consistency are more pronounced than the variations in planes. In weakly textured regions, where photometric consistency loses reliability, the prior plane information will play a major role in the hypothesis updating.

\subsection{Geometric Consistency}
Due to the presence of weakly textured regions, where even incorrect depth assumptions can lead to patches matching sufficiently similar regions, the issue of ambiguous matching arises. ACMP effectively addresses the problem of photometric consistency by imposing geometric consistency constraints on the aggregation cost. This constraint serves as a measure of the quality of depth estimation, mitigating the challenges posed by ambiguous matching in weakly textured regions.
\par
Given the depth $D_{i}(p)$ corresponding to pixel $p$ in image $I_{i}$, and the camera parameter $P_{i}=\{ M_{i}|\text{p}_{i,4} \}$, its corresponding back-projected 3D point $X_{i}(p)$ can be expressed by the following equation:
\begin{eqnarray}
X_{i}(p)=M^{-1}_{i}\cdot (D_{i}(p)\cdot p-\text{p}_{i,4}),
\end{eqnarray}
\par
Then, we can calculate the reprojection geometric consistency error between the reference image $X_{ref}$ and the source image at pixel $p$ as:
\begin{eqnarray}
Cost_{geo}=\min(\parallel P_{ref}\cdot X_{src}(P_{src}\cdot X_{ref}(p)-p)\parallel ,\tau _{geo}), 
\end{eqnarray}
where $\tau _{geo}$ is a truncation threshold to robustify the reprojection error against occlusions.

\section{Algorithm}
In this section, we present the process of our proposed algorithm. To address the issue of Delaunay triangulation segmentation fragmentation, we introduce SAM Planar Prior to accurately segment the weakly-textured regions, and generate more complete prior planar information (see Section \ref{SAM Planar Prior}). Furthermore, while ACMP only considers local consistency in the prior assisted update. To more effectively integrate prior information into the depth estimation process, we enhance geometric consistency by introducing epipolar line constraints (see Section \ref{Geometric Consistency}), which we then combine with global information aggregation (see Section \ref{Global Information Aggregation Cost}) via probability graph model (see Section \ref{Update of the Hypothesis}).
\par
As illustrated in \autoref{fig3}, in the first stage, we utilize ACMP algorithm to obtain raw depth, then the plane prior candidates composed of Delaunay triangulation prior and SAM Planar Prior (SPP) are generated. In the second stage, embedding prior candidates set into the update of depth estimation employing global information aggregation cost function (GIA), derived from the probabilistic graphical model, along with Geometric Consistency For Epipolar Constraints (CGEC). Finally, we obtain the final depth estimation after PatchMatch MVS with geometric consistency.
\begin{figure*}[ht]
  \centering
  \begin{minipage}{1\linewidth}  
    \centering
    \subfigure[Images]{\includegraphics[width=0.3\linewidth]{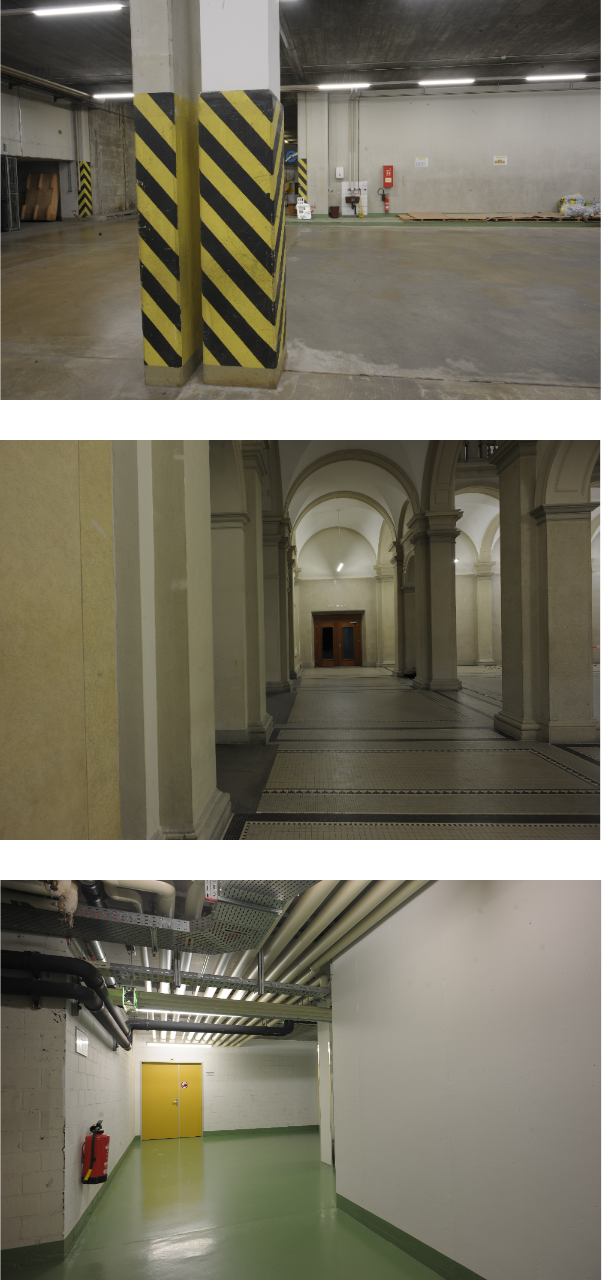}
    \label{fig7a}}
    \subfigure[SAM Segment]{\includegraphics[width=0.3\linewidth]{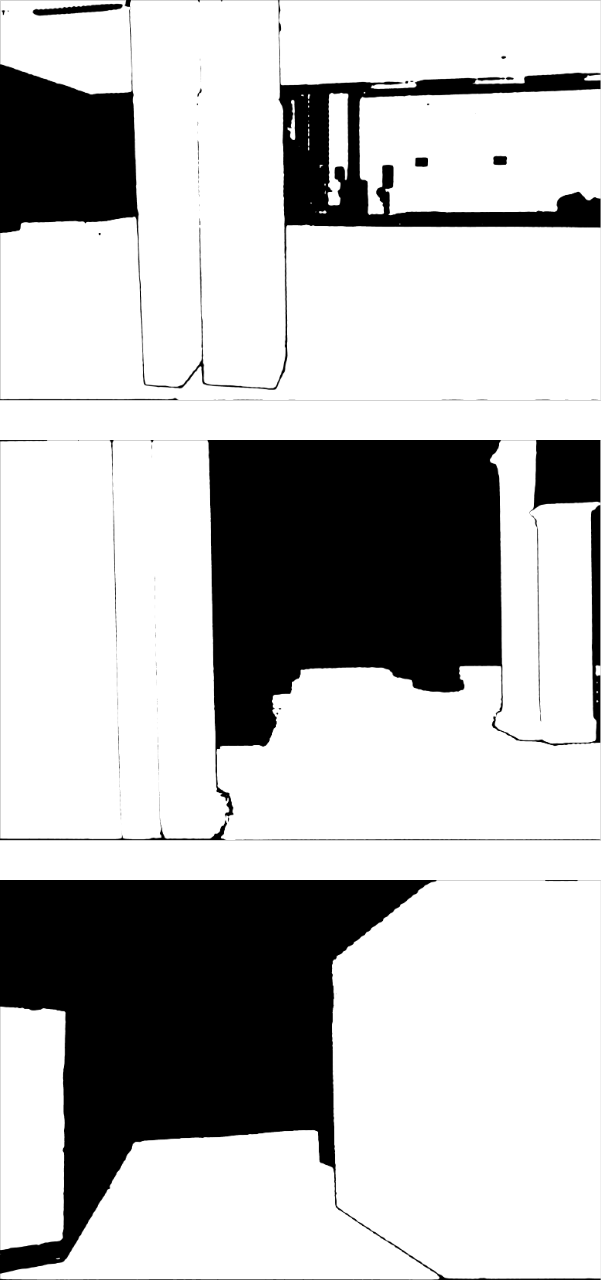}
    \label{fig7b}}
    \subfigure[Tri Segment]{\includegraphics[width=0.3\linewidth]{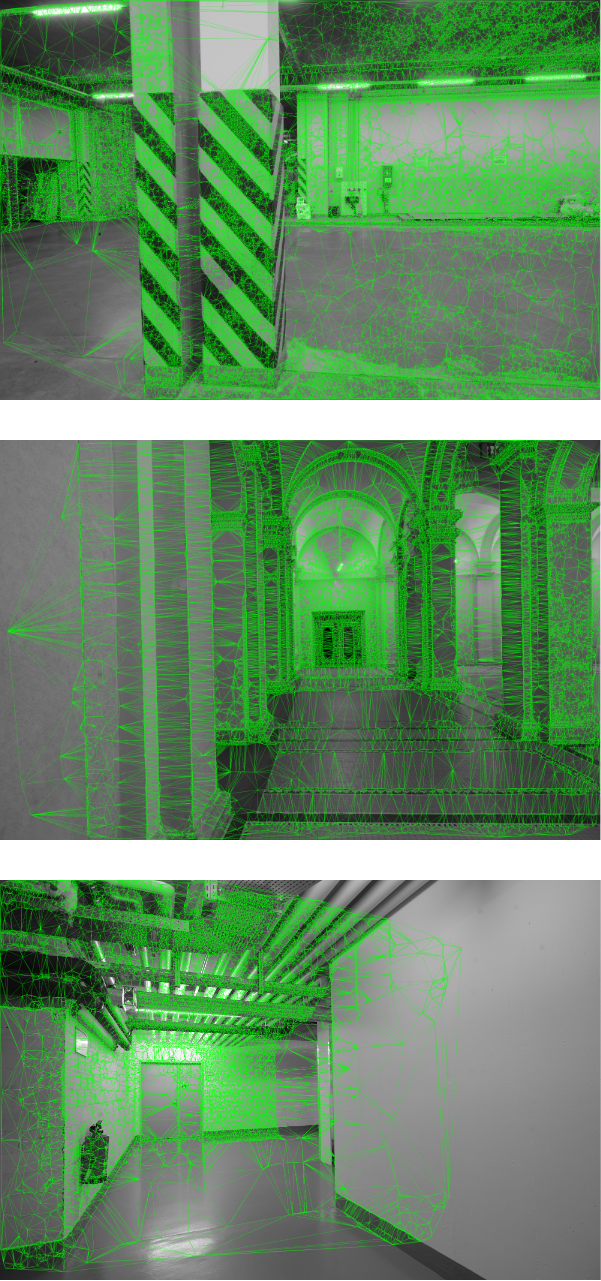}
    \label{fig7c}}
  \end{minipage}
  \caption{(a) ETH3D high-resolution images, (b) Segment mask obtained through SAM, (c) Triangulation segmentation.}
  \label{fig7}
\end{figure*}
\subsection{SAM Planar Prior}
\label{SAM Planar Prior}
In the previous sections, we explore the challenges associated with weakly-textured regions in 3D reconstruction and the limitations of traditional methods in addressing these challenges. To overcome these limitations, our approach leverages advanced segmentation techniques to accurately identify and model weakly-textured regions. One of the key components of our approach is the integration of the Segment Anything Model, a state-of-the-art segmentation model. SAM plays a crucial role in our pipeline by providing precise segmentation masks, which are essential for effective plane fitting in weakly-textured regions. In the following section, In the next section, we dive into how SAM can be utilized to construct the prior plane, explaining how it can improve the accuracy and reliability of our 3D reconstruction process, particularly in challenging environments.
\par
\textbf{Segment Anything}: We employ the SAM to perform the segmentation task of weakly-textured regions in images. SAM is composed of an image encoder, a prompt encoder, and a mask decoder. The image encoder is based on a visually improved transformer pre-trained with Masked Autoencoders, primarily tasked with converting input images into embedded forms of input tensors. The prompt encoder transforms prompt information (points, boxes, and text) into a representation form of positional encoding and embeds these prompts by summing them with the output of the image encoder through learned mechanisms. Finally, the mask decoder efficiently maps image embeddings and prompt embeddings to masks, computing the foreground probability for each position in the image. Overall, SAM is a promptable foundational model trained on over a billion masks from 11 million images for promptable segmentation tasks, achieving powerful zero-shot generalization.
\par
Our method is to perform plane fitting under the guidance of masks, the accuracy of segment masks is critical for the effectiveness of plane fitting and the overall quality of the 3D reconstruction, particularly in weakly-textured regions. Inaccurate masks can lead to poor plane fitting, incorrect modeling of weak texture areas, and propagation of errors through subsequent reconstruction steps, ultimately degrading the quality of the final 3D model. By using a point within the weakly-textured region as the prompt, SAM can generate a accurate mask image specific to that area that overcome issues of over-segmentation or fragmentation, which ensures more precise plane fitting and better modeling of weakly-textured regions. We semi-automatically calibrate regions of interest in the scene (such as the ground or walls) as prompts. This guides SAM to more accurately segment weakly-textured regions in the images, ensuring robustness, particularly during prior plane generation in subsequent tasks. \autoref{fig7}(b) showcases the accurate segmentation results, preserving the majority of large weakly-textured regions in the scene. The obtained masks are then applied for plane fitting in the following sections.
\par
\textbf{Prior Plane Generation}: The acquired masks guide the RANSAC algorithm in defining the fitting range, enabling subsequent plane fitting on the filtered 3D points from ACMP. This approach leads to fitted planes with a high level of completeness.
\par
Firstly, we utilize the ACMP algorithm to compute rough depth maps and normal vector maps for each image. We obtain the pixel coordinates $p_{i}(x_{i},y_{i})$ within a specific segmentation regions sequentially from the segmentation results. These coordinates are projected to camera coordinates using the following equation to derive the 3D point $X(x_i^c,y_i^c,z_i^c)$:
\begin{eqnarray}
\begin{bmatrix}x_i^c \\y_i^c \\z_i^c\end{bmatrix} =         D(p_{i})\times  
    \begin{bmatrix} \frac{x_{i}-c_{x}}{f_{x}}  \\\frac{y_{i}-c_{y}}{f_{y}} 
    \\1\end{bmatrix}
,
\end{eqnarray}
where $D(p_{i})$ is the depth corresponding to pixel $p_{i}$.
\par
Then, we filter 3D points within fitting regions using Principal Component Analysis (PCA) to eliminate uneven points, ensuring efficient and high-quality plane fitting. For each 3D point $X_i(x_i,y_i,z_i)$ in the fitting regions, the covariance matrix is calculated using \autoref{eq9}:
\begin{eqnarray}
Cov=\frac{1}{N} \sum_{i=1}^{N}(X_{i}-\bar{X})(X_{i}-\bar{X})^{T},
\label{eq9}
\end{eqnarray}
where $ N $ is the number of points in the fitted regions and $ \bar{X} $ is the average position of all points in the regions. Next, the eigenvalues of the Cov matrix $ \lambda _{1}>\lambda _{2}>\lambda _{3} $ are calculated and the curvature of the regions of this fitting plane is calculated according to \autoref{eq10}:
\begin{eqnarray}
Cur=\frac{\lambda _{1}}{\lambda _{1}+\lambda _{2}+\lambda _{3}}. 
\label{eq10}
\end{eqnarray}
The curvature values $Cur$ range from 0 to 1, representing the curvature of the point cloud. A value closer to 1 indicates a flatter point cloud, while a value closer to 0 suggests higher curvature. In our assumption, the 3D points in the fitted plane should adhere as much as possible to the planar characteristics within their local neighborhood. Therefore, we retain points in the fitting regions that are relatively flat by setting a threshold $\tau_{\lambda}$.  
\par
The RANSAC algorithm iteratively fits the filtered 3D point set to generate SAM-based plane priors. This involves randomly selecting and iterating data points to find the most consistent model. The final result is projected onto a 2D plane. From \autoref{fig7} and \autoref{fig8}, our method accurately segments most large weakly-textured regions, leveraging planarity for smooth and complete planes. In contrast, triangulation struggles to partition extensive planes, causing fragmentation. Moreover, triangulation-based priors are vulnerable to errors in depth estimation, culminating in the generation of inaccurate plane priors. SAM-based priors prove more robust.

\begin{figure}
    \centering
    \includegraphics[width=8cm,height=6cm]{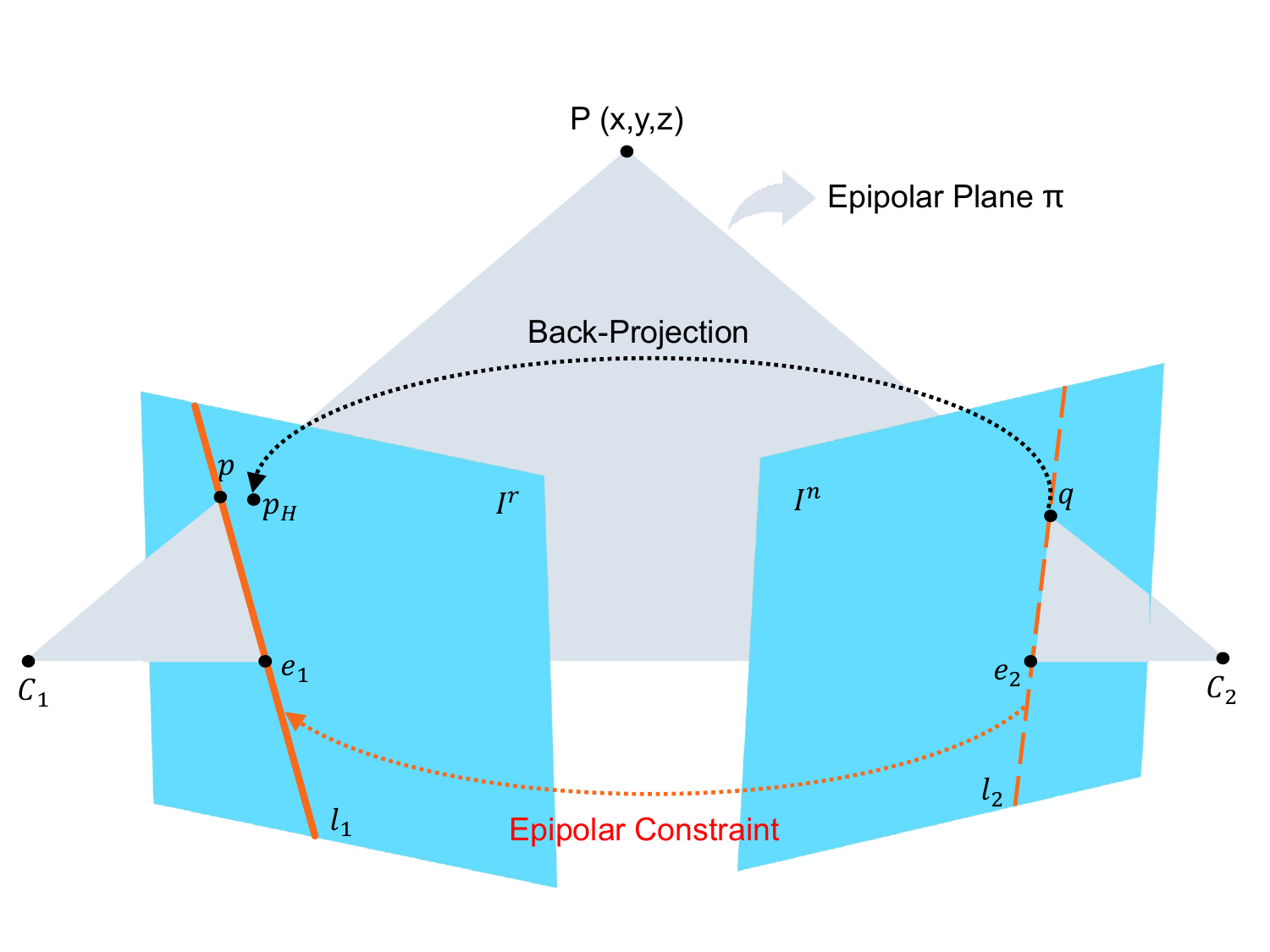}
    \caption{Epipolar Constraints. The camera centers of the left and right images are $C_1$ and $C_2$, respectively. $P$ is a coordinate in the world coordinate system, which represents a point in the scene. $C_1$, $C_2$, and $P$ create an epipolar plane. The epipolar lines $l_1$ and $l_2$ are the lines of intersection between the epipolar plane and the two image planes $I_r$ and $I_s$.}
    \label{fig4}
\end{figure}

\subsection{Geometric Consistency For Epipolar Line Constraints}
\label{Geometric Consistency}
In the previous sections, we introduced the construction of the prior plane to obtain better prior depth information in more textured regions. However, there are problems in how to effectively utilize the prior information. Due to the matching ambiguity problem caused by photometric consistency, the correct prior information can not be used. Therefore, we introduce geometric consistency to constrain the influence of the matching ambiguity problem.
\par
Geometric consistency represents the relationship between projecting a pixel from the reference image to the source image and then reprojecting it. In traditional MVS algorithms, geometric consistency errors are limited by a truncation threshold, which can lead to challenges such as loss of accuracy, local structure neglect, and noise sensitivity. 
\par
To cope with this challenge and accurately describe the 3D positional relationship between matching points, \cite{r51} introduce geometric consistency based on epipolar constraints into the matching cost function by utilizing an adaptive threshold that takes into account whether a point
lies on an epipolar line or not. Assuming pixel $q$ is the matched pixel of pixel $p(x_{1},y_{1})$ on the source image $I_s$, and pixel $p_H(x_{2},y_{2})$ is the pixel where pixel $q$ is back-projected to the reference image $I_r$, and $l_1$ is the epipolar line of pixel $q$ on the reference image $I_s$. Their positional relationship is illustrated in \autoref{fig4}. If pixels $p$ and $q$ are correctly matched, then pixel $p_H$ should lie on the epipolar line $l_1$ $(ax+by+c=0)$ and coincide with pixel $p$. Therefore, we obtain an adaptive threshold:
\begin{eqnarray}
\mathfrak{D}_{sta}=\begin{cases}
\frac{\frac{c}{a}(x_{2}-x_{1})-\frac{c}{a}(y_{2}-y_{1}) }{\sqrt{(\frac{c}{a})^{2}+(\frac{c}{b})^{2}+\mathfrak{D}_{H}} } \mathfrak{D}_{H} & \text{ if } \mathfrak{D}_{p}=0 \\
 \quad  \quad  \quad  \mathfrak{D}_{p} & \text{ if } \mathfrak{D}_{p}\ne 0
\end{cases},
\end{eqnarray}
where $\mathfrak{D}_p$ represents the distance from pixel $p$ to the epipolar line $l_1$, and $\mathfrak{D}_H$ represents the distance between pixel $p$ and pixel $p_H$.
\par
The matching cost for geometric consistency is calculated as follows:
\begin{eqnarray}
Cost_{geo}=\begin{cases}
 \frac{\mathfrak{D}_{H}}{\mathfrak{D}_{sta}}+(1-V_{sim})  & \text{ if } \mathfrak{D}_{H} < \omega_{geo}\mathfrak{D}_{sta} \\
\quad \quad  \quad   2 & \text{ if }\mathfrak{D}_{H}\ge \omega_{geo}\mathfrak{D}_{sta}

\end{cases},
\end{eqnarray}
where $\omega _{geo}$ is the range parameter used to determine the pixel $p_H$. $V_{sim}$ is the similarity of the normal vectors $\vec{n}_p,\vec{n}_q$ of pixel $p$ and pixel $q$. It is calculated as follows:
\begin{eqnarray}
V_{sim}=\frac{\vec{n}_{p}\cdot\vec{n}_{q}  }{\left | \vec{n}_{p} \right | \left | \vec{n}_{q} \right | }.
\end{eqnarray}

\subsection{Global Information Aggregation Cost}
\label{Global Information Aggregation Cost}
ACMP only considers local consistency in the prior assisted update. To more effectively integrate prior information into the depth estimation process, we draw inspiration from the SGM \citep{r28}, where the central idea is to integrate depth information of individual pixels into a globally consistent result. Motivated by this idea, we introduce the concept of cost aggregation. 
\par
\begin{figure}
    \centering
    \includegraphics[width=5cm,height=4cm]{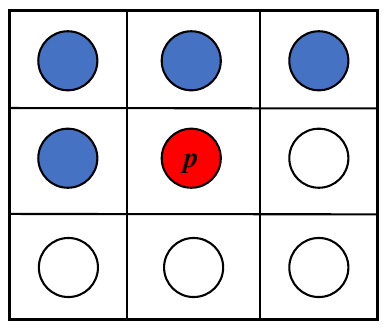}
    \caption{The selection of domain pixels in the cost of global information aggregation. $p$ is the current pixel that need to be aggregated, the blue part represents the selected domain pixels of $p$, while the others white part indicates the pixels that have not been aggregated.}
    \label{fig5}
\end{figure}

We adopt the top-to-bottom and left-to-right patterns to enable sequential aggregation and updating of the cost function. This ensures that each pixel gathers information from its recently updated neighbors, thereby enhancing the accuracy and consistency of depth estimation. The utilization of four neighboring pixels provides sufficient contextual information for smoothing the cost function while also ensuring computational feasibility and efficiency. By establishing an aggregation function, we consider not only the matching relationship between the current image and its neighboring images but also the overall information of the current image. This aggregation function consists of two parts: the matching cost of the current pixel and the aggregation cost of multiple neighboring pixels. This implies that the aggregation cost of pixels is not limited to the local neighborhood but is determined by comparing global consistency. The final depth estimation depends not only on local features but also on the influence of the overall scene structure, thereby enhancing the algorithm's robustness and accuracy. The formula for calculating the global information aggregation cost is as follows:
\begin{eqnarray}
L_{agg}=Cost_{match}+L_{smooth},
\label{eq11}
\end{eqnarray}
\begin{eqnarray}
L_{smooth}=\frac{\min(L_{1},L_{2},L_{3},L_{4})}{L_{1}+L_{2}+L_{3}+L_{4}},
\label{eq12}
\end{eqnarray}
where $Cost_{match}$ represents the current match cost of pixel $p$, $L_{smooth}$ is the global aggregation cost smoothing term, and $\left\{ L_{1}, L_{2}, L_{3}, L_{4} \right\}$ denote the global information aggregation cost of pixel $p$'s neighboring pixels. As shown in the \autoref{fig5}, the blue regions represents the neighborhood pixel of pixel $p$, whose global information aggregation cost is also calculated by the \autoref{eq11}. Hence, \(L_{agg}\) is acquired through multiple nested accumulations, signifying the matching cost for the entire image.
\par
From the \autoref{eq11}, it can be seen that this global information aggregation cost combines both local and global information. This means that it can deal not only with regions with obvious local features but also with regions showing stronger global consistency, such as flat surfaces or weakly-textured regions. In complex scenes, factors such as illumination changes and occlusion may lead to difficulties in local matching, but the global information can strengthen the stability of the matching on the whole and provide more reliable depth estimation results.

\begin{figure}
    \centering
    \includegraphics[width=6cm]{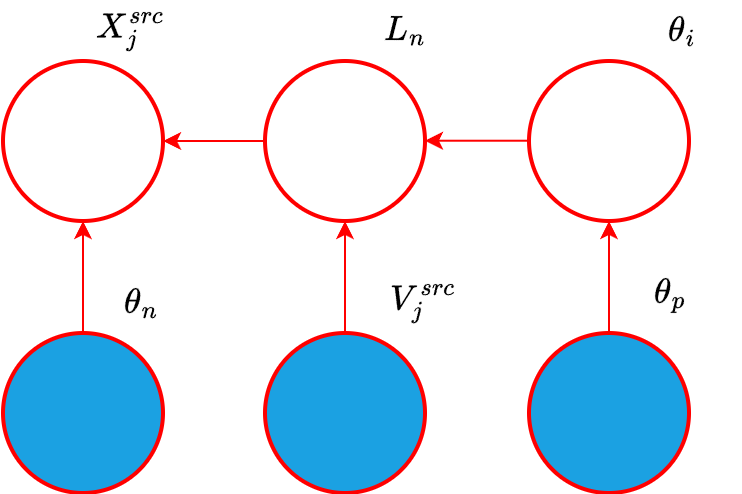}
    \caption{Probability graph model. $\theta_i$ is a plane candidate. Given planar prior $\theta_p$ with it's neighboor prior $\theta_n$, the gobal information aggreation $L_n$ and the observation $X_{i}^{src}$ on source images with the visibility information $V^{src}_i$, the updated hypotheses $\theta_i^*$ is inferred.}
    \label{fig6}
\end{figure}

\begin{figure*}[ht]
  \centering
  \subfigure[Images]{\includegraphics[width=1.3in]{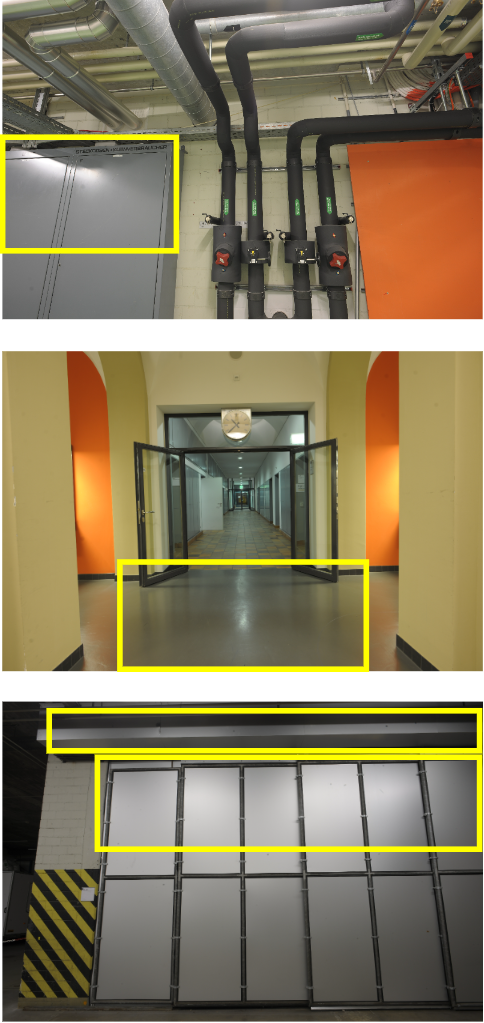}
  \label{images}}
  \hspace{-1mm}
  \subfigure[Raw Depth]{\includegraphics[width=1.3in]{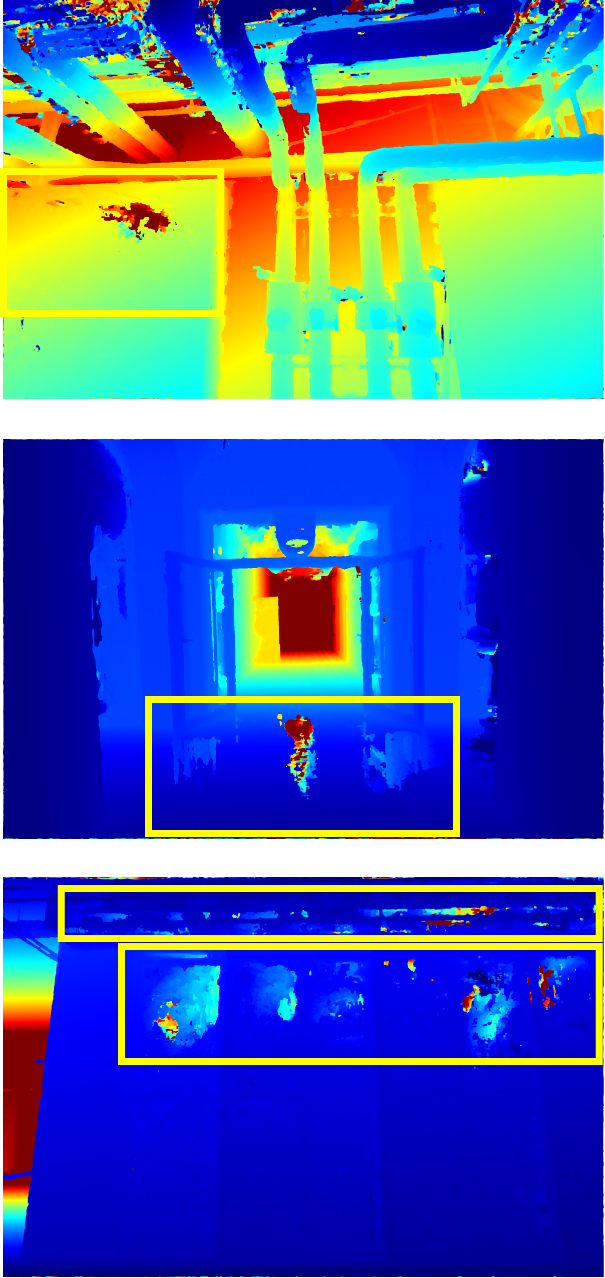}
  \label{raw}}
  \hspace{-1mm}
  \subfigure[SAM-based Prior]{\includegraphics[width=1.3in]{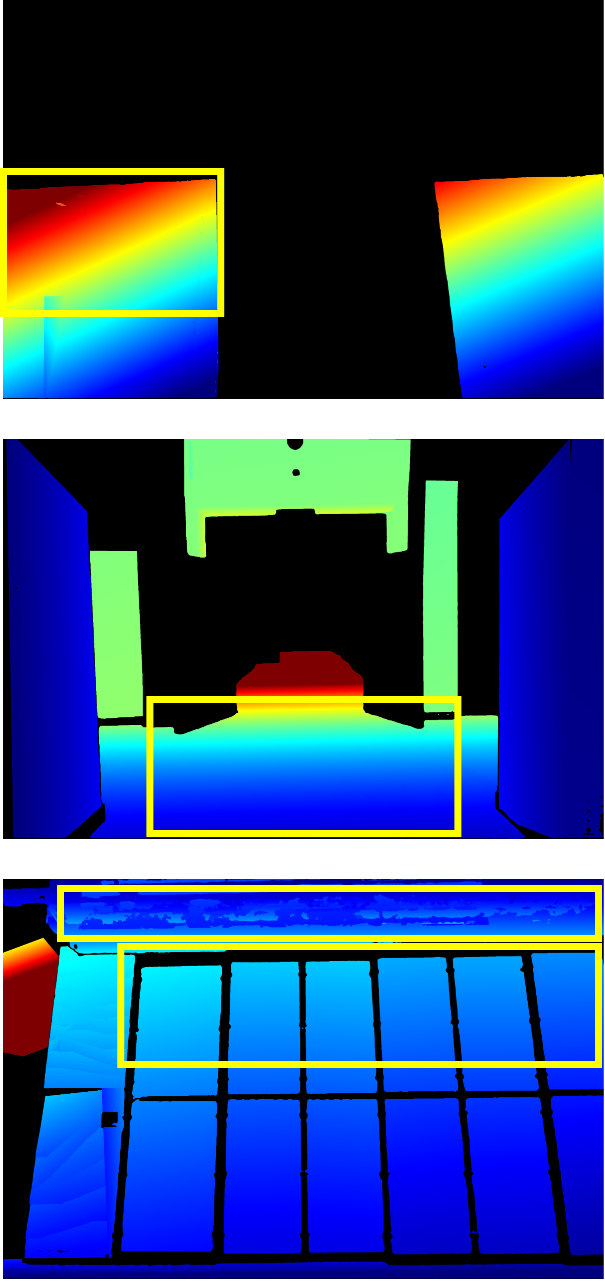}
  \label{fig8sam}}
  \hspace{-1mm}
  \subfigure[Tri Prior]{\includegraphics[width=1.3in]{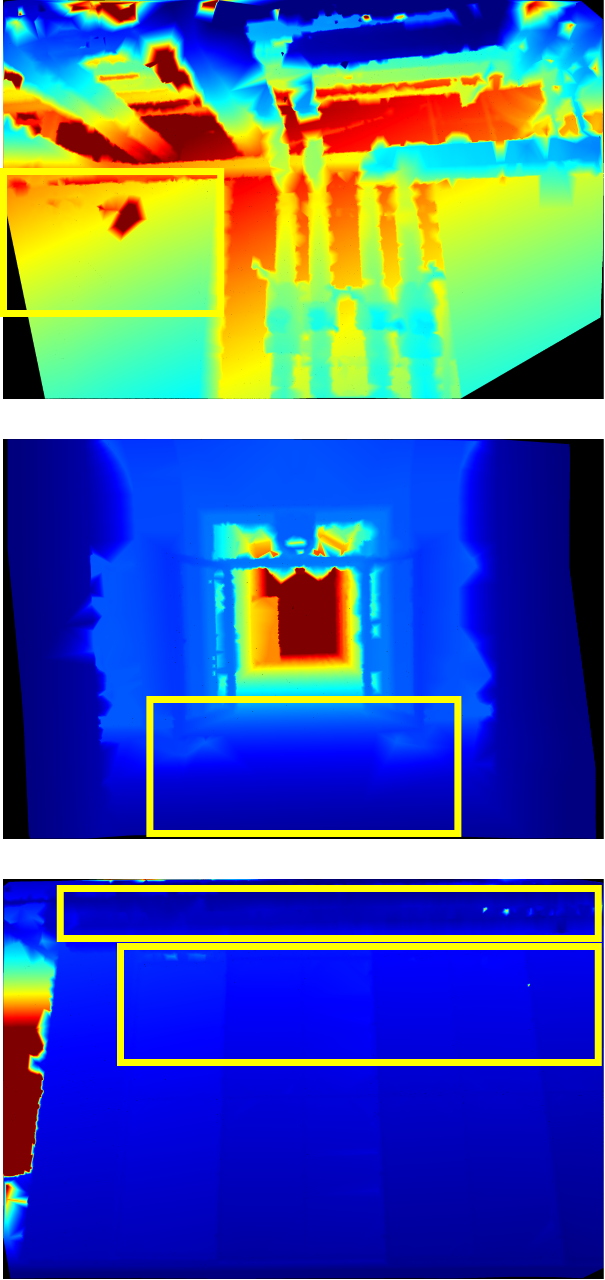}}
  \label{fig8tri_p}
  \hspace{-1mm}
  \subfigure[Updated Depth]{\includegraphics[width=1.3in]{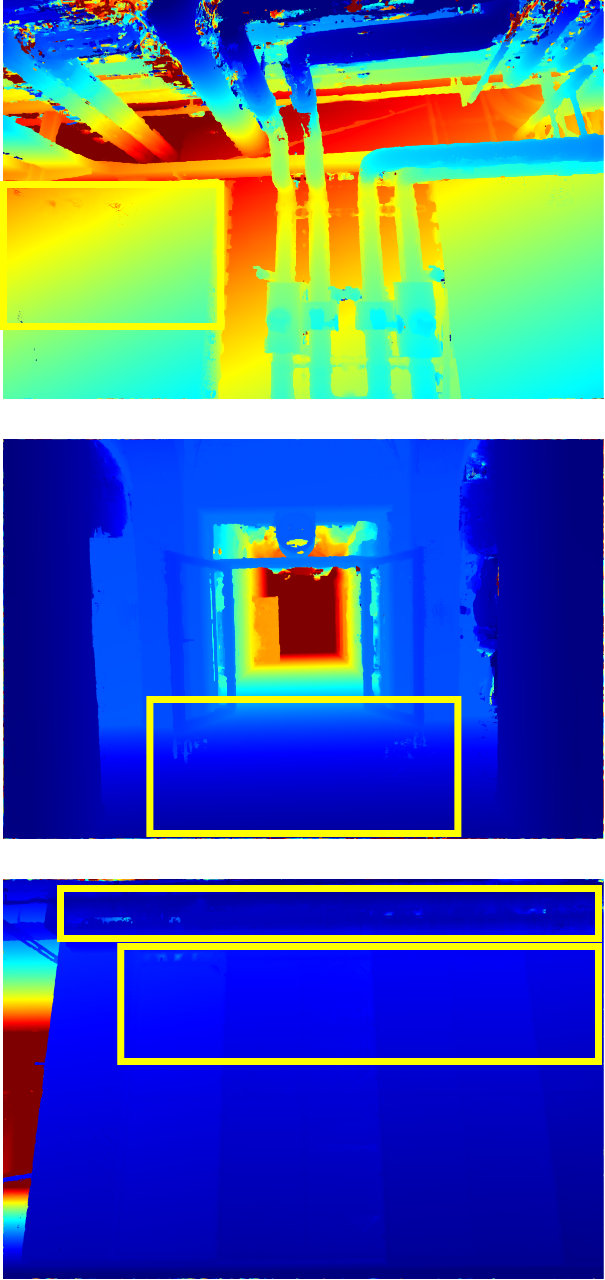}}
  \label{fig8agg}
  \caption{(a) ETH3D high-resolution images, (b) Raw depth maps obtained through ACMP algorithm, (c) Prior depth maps obtained through SAM, (d) Prior depth maps obtained through triangulation, (e) Depth maps that are updated with our method. Some challenging regions are shown in yellow boxes.}
  \label{fig8}
\end{figure*}

\subsection{Update of the Hypothesis}
\label{Update of the Hypothesis}
As previously discussed, with the planar prior, the depth estimates for weakly-textured regions can be better approximated. However, to further refine the depth estimates and incorporate prior knowledge effectively, we must consider a more holistic approach to cost aggregation. In this section, we construct the aggregation cost function employing probabilistic graph model. This transforms the task of finding the optimal hypothesis for each pixel into a problem of minimizing the aggregation cost function. Consequently, the plane prior is incorporated into the depth estimation update process in the form of a prior constraint.
\par
Given prior hypotheses and the depth and normal vector estimates of neighboring pixels, we utilize a probabilistic graph model to obtain a multi-view cost aggregation function for the prior information. This allows the prior information to be involved in the depth estimation update. The probabilistic graph model we constructed is illustrated in \autoref{fig6}, with its joint probability defined as:
\begin{eqnarray}
\begin{split}
& P(X_{j}^{src},V_{j}^{src},L_{n},\theta_{i},\theta_{p},\theta_{n})\propto \\
& P(X_{j}^{src}\mid L_{n},\theta_{n})P(L_{n}\mid\theta_{i},V_{j}^{src})P(\theta_{i}\mid \theta_{p}).
\end{split}
\label{eq13}
\end{eqnarray}
Similar to ACMP, we define the prior plane hypothesis as $\theta_{p}$, neighborhood pixel hypothesis $\theta_{n}$, candidate plane hypothesis $\theta_{i}$, observations on the source image $X_{j}^{src}$, visibility information $V_{j}^{src}$, and global information aggregation cost $L_{n}$ in the neighborhood. Our goal is to derive the optimal hypothesis $\theta_{i}^{*}$ based on the probability graph model. That is, to obtain the optimal hypothesis $\theta_{i}^{*}$ by calculating the maximum posteriori hypothesis of the plane hypothesis, whose maximum posteriori estimate is as follows:
\begin{eqnarray}
\theta_{i}^{*} = \arg \max P(\theta_{i}\mid X_{j}^{src},V_{j}^{src},\theta_{p},\theta_{n},L_{n}).
\label{eq14}
\end{eqnarray}
The above posterior probabilities can be decomposed as follows:
\begin{eqnarray}
\label{eq15}
\begin{split}
& P(\theta_{i}\mid X_{j}^{src},V_{j}^{src},\theta_{p},\theta_{n},L_{n})\propto \\
& P(X_{j}^{src}\mid L_{n},\theta _{n})P(L_{n}\mid\theta_{i},V_{j}^{src})P(\theta_{i}\mid \theta_{p}).
\end{split}
\label{eq16}
\end{eqnarray}
According to photometric consistency \citep{r5, r34}, the cost function is defined as:
\begin{eqnarray}
Cost_{ph}(\theta_i)=\frac{\sum_jw_j\cdot m_{i,j}}{\sum_jw_j},
\end{eqnarray}
where \(m_{i,j}\) represents the photometric consistency cost between the reference patch \(i\) and the corresponding patch \(j\) in the source image, and \(w_{j}\) denotes the view selection weight for the source image.
We define the likelihood function as:
\begin{eqnarray}
P(L_{n}\mid\theta_{i},V_{j}^{src})=e^{-Cost_{ph}}.
\label{eq17}
\end{eqnarray}
We introduce geometric consistency based on epipolar constraints in the likelihood function. The planar prior is defined as:
\begin{eqnarray}
P(\theta_{i}\mid\theta_{p})=e^{-\alpha_{geo}\cdot Cost_{geo}},
\label{eq18}
\end{eqnarray}
Where $a_{geo}$ is the weight parameter of the geometric consistency cost. The combination of prior information and geometric consistency effectively addresses the matching ambiguity in photometric consistency.

\par
Since the above matching cost calculation only considers local correlation, it is susceptible to interference from noise. To achieve a better matching effect, we design a likelihood function incorporating cost aggregation. The aggregated surrogate value more accurately reflects the correlation between pixels. The likelihood function, after the addition of cost aggregation, is as follows:
\begin{eqnarray}
P(X_{j}^{src}\mid L_{n},\theta_{n})=e^{-L_{smooth}},
\label{eq19}
\end{eqnarray}
where $L_{j}$ is the cost aggregation function within the neighborhood of the current pixel. This likelihood function ensures both local and global consistency. The aim is to make the depth and normal vector estimates of the current pixel most consistent on a local scale by selecting the smallest aggregation cost function for neighboring pixels. If the depth and normal vector hypotheses of a pixel are consistent with those of the surrounding pixels, they have a smaller aggregation cost and are more likely to be selected as the final depth and normal vector. Simultaneously, selecting the optimal hypotheses over the information of the entire image ensures global consistency. 
\par
To obtain the global information aggregation cost for the current pixel $p$, we substitute \eqrangeref{eq15}{eq19} into \autoref{eq14} and use the negative logarithmic algorithm, yielding:
\begin{eqnarray}
L_{agg}=Cost_{ph}+\alpha _{geo}\cdot Cost_{geo}+L_{smooth}.
\label{eq20}
\end{eqnarray}
The cost aggregation function integrates a global optimization strategy based on photometric and geometric consistency, along with neighborhood consistency. Built upon these consistency cues, it selects depth and normal vector hypotheses with the minimum aggregation cost. This enhances the accuracy of overall matching, avoids falling into local optima, and improves the accuracy and robustness of depth estimation.
\par
In \autoref{eq20}, we calculate the global information aggregation costs $L_{agg}^{SAM}$, $L_{agg}^{TRI}$, and $L_{agg}^{RAW}$ for the SAM-based prior plane, triangulation prior plane, and ACMP raw estimation plane, respectively. Finally, we update the current hypothesis employing the probability graph model, i.e., by determining the optimal plane hypothesis $\theta ^{*}=(D^{*},\vec{n}^{*})$ for the ith pixel in the current image through the following formula: 
\begin{eqnarray}
(D^{*},\vec{n}^{*})=\arg \min \left\{\begin{matrix}&L_{agg}^{{\scriptsize SAM}}\\ \\& L_{agg}^{{\scriptsize TRI}}+P_{1}\\ \\&L_{agg}^{{\scriptsize RAW} }+P_{2}\end{matrix}\right.
,
\end{eqnarray}
where $P_1$ and $P_2$ are penalty terms for the triangulation prior and ACMP raw estimation planes. Introducing penalty terms allows for a balance between different priors, adapting better to the characteristics of various scenes. In \autoref{fig8}(e), the updated depth estimation is showcased, revealing our method's precise estimation of depth in weakly-textured regions. This approach optimizes the process by amalgamating information from three distinct sources, yielding a more dependable and globally consistent depth estimation result. 
\par
In summary, our method strategically leverages the strengths of diverse depth estimation techniques, integrating information from all three sources through a probabilistic graph model. This integration results in a depth estimation that is not only reliable but also globally consistent.

\section{Experiments}
In this section, we compare our method with several state-of-the-art PatchMatch-based MVS techniques, including COLMAP, ACMP, ACMMP, TAPA-MVS, QAPM \\ \citep{r31}, TSAR-MVS \citep{r33} and HC-MVS \citep{r51}. We assessed the performance of our method on multiple high-resolution datasets, conducting evaluations through both quantitative and qualitative measures. All of our experiments are conducted on a Linux system equipped with an Intel Xeon Platinum CPU and an NVIDIA A100 GPU.

\begin{figure*}[ht]
  \centering
  \includegraphics[width=\linewidth]{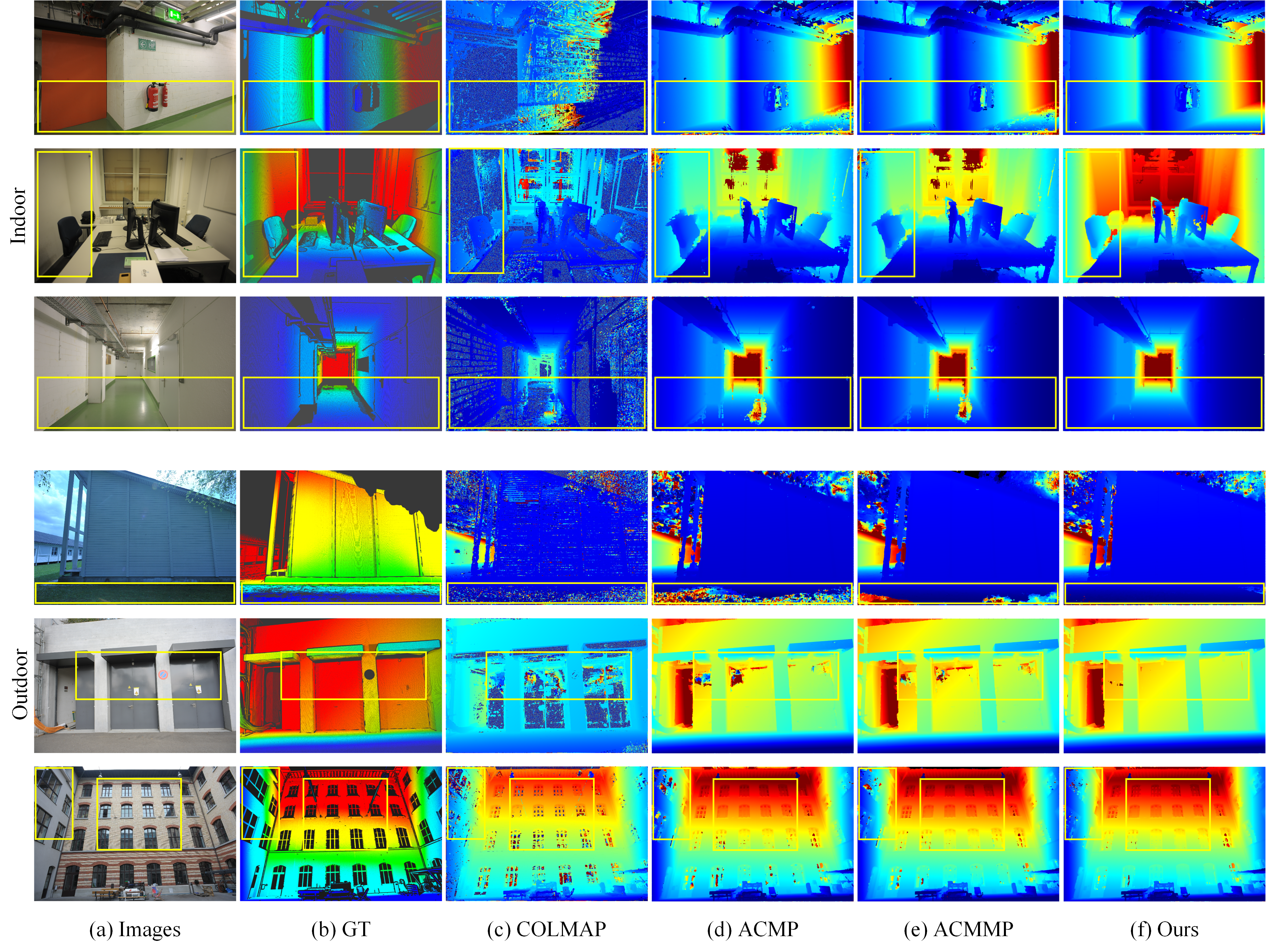}
  \caption{Some indoor and outdoor high-resolution multi-view datasets from the ETH3D benchmark (pipes, office, terrains, meadow, electro, courtyard), various algorithms are used to compare qualitative depth maps: (c) COLMAP; (f) ACMP; (e) ACMMP; (f) Ours. Black pixels in the GT indicate no data. Some challenging regions are shown in yellow boxes.}
  \label{fig9}
\end{figure*}

\begin{figure*}[p]
  \centering
  \includegraphics[width=0.93\linewidth, height=0.9\textheight]{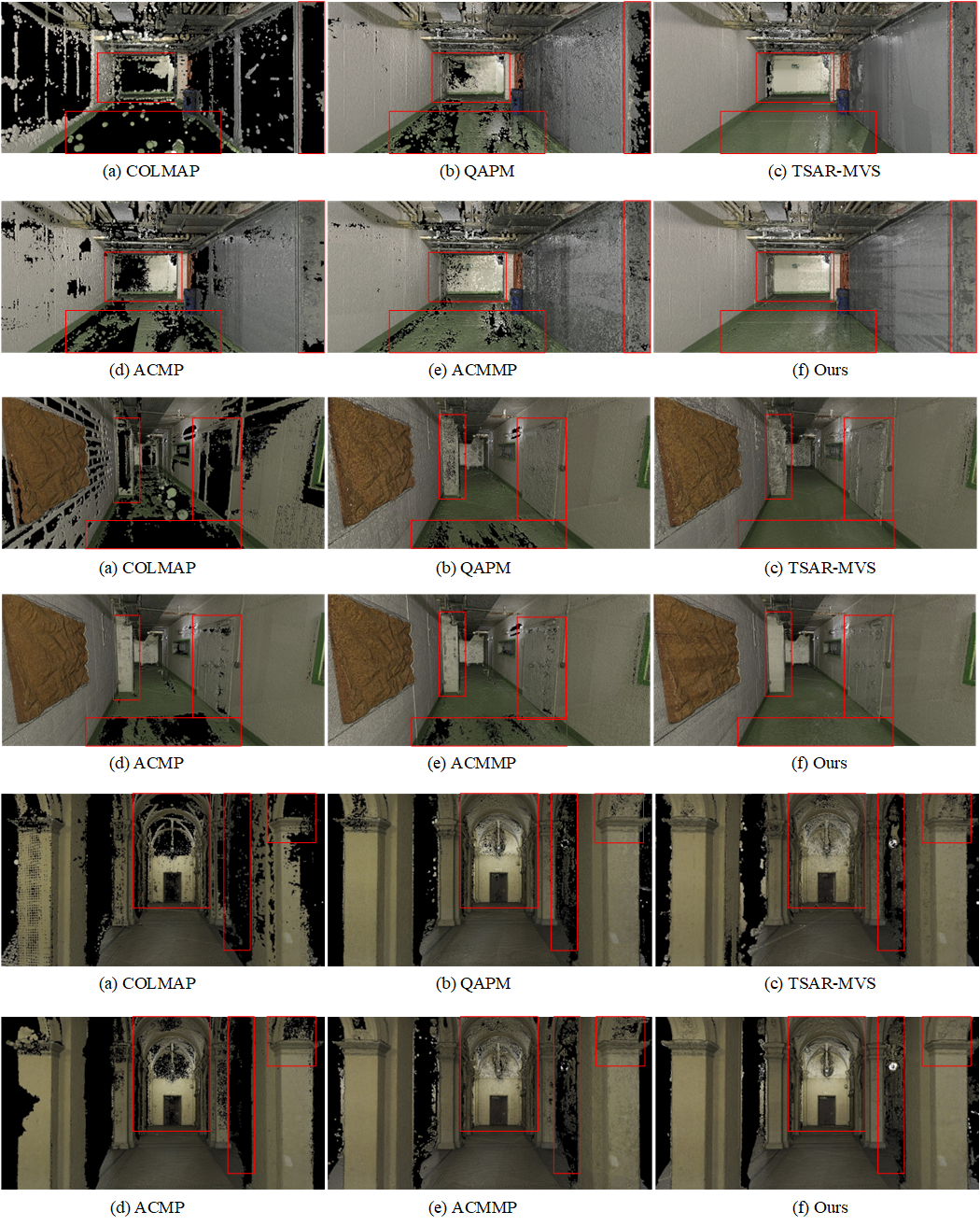}
  \caption{Qualitatively compare point clouds from various algorithms on some images from the ETH3D benchmark dataset (pipes, terrains, relief): (a) COLMAP; (b) QAPM; (c) TSAR-MVS; (d) ACMP; (e) ACMMP; (f) Ours. Challenging regions are marked in red boxes, and black pixels in the point cloud represent areas with no data.}
  \label{fig10}
\end{figure*}

\begin{figure*}[p]
  \centering
 \includegraphics[width=0.93\linewidth, height=0.9\textheight]{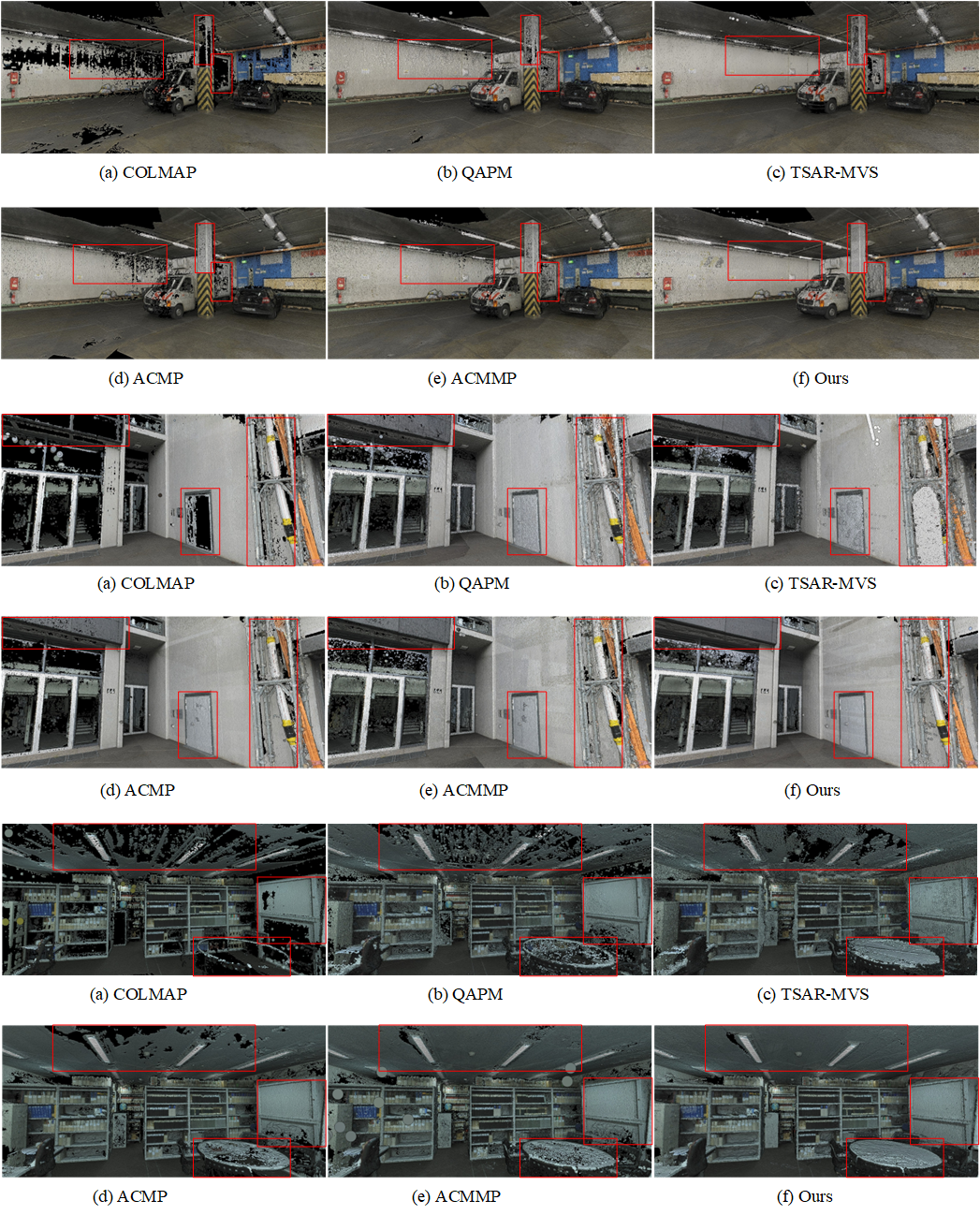}
  \caption{Qualitatively compare point clouds from various algorithms on some images from the ETH3D benchmark dataset (delivery, electro, kicker): (a) COLMAP; (b) QAPM; (c) TSAR-MVS; (d) ACMP; (e) ACMMP; (f) Ours. Challenging regions are marked in red boxes, and black pixels in the point cloud represent areas with no data.}
  \label{fig11}
\end{figure*}

\subsection{Datasets and Parameters Settings}
\textbf{Datasets}: To verify the reliability and generalization of our method, we employ the ETH3D benchmark dataset \citep{r30} as our benchmark dataset. For our experiments and analysis, we focus on the high-resolution dataset to evaluate the performance of our model in handling large-scale scenes with high resolution, originally sized at 6048×4032. This dataset consists of 13 training sets and 12 test sets. The Ground Truth (GT) is exclusively provided by the training sets. The GT data are acquired through high-precision lidar, and thus can be considered as the true values for the scene depth. Additionally, we conducted a qualitative evaluation of the reconstructed point clouds on the UDD5 \citep{r32} and SenseFly-University Perception datasets, with original image resolutions of 4000×3000 and 6000×4000, respectively. These datasets are collected from remote sensing images captured using unmanned aerial vehicles (UAVs). 
\par
Furthermore, we also test the performance of our algorithm on public datasets provided by \href{https://www.agisoft.com/}{agisoft} and real scenes captured by ourselves. The resolution of the original images of agisoft dataset is 4368 x 2912, respectively. The real scene is captured by a NIKON Z50 camera and consists of 42 images with the resolution of 5568 x 3712.
\par
\textbf{Parameter Settings}: Our method is implemented based on C++11 and CUDA11.8. The parameter values are set as $\{\tau_{\lambda},\omega_{geo},\alpha_{geo},P_1,P_2\}=\{0.5,5,0.1,0.2,0.66\}$. $\tau_{\lambda}$ filters out non-flat noise points in the plane fitting process. $\omega_{geo}$ serves as the threshold for prior geometric consistency, incorporating epipolar line constraints. $\alpha_{geo}$ represents the weight of geometric consistency. $P_1$ and $P_2$ act as penalties for triangulation prior and raw depth, respectively. Triangulation is particularly effective in small, planar, weakly-textured regions compared to SAM-based prior. However, its small-area planar priors are more reliable than raw depth, leading to values of 0.2 and 0.66, respectively. It's noteworthy that, for consistency with other models, we set the image resolution during experiments to 3200×2016. 

\subsection{Depth Map Evaluation}
The ETH3D datasets features architectural structures with numerous weakly-textured regions. \autoref{fig9} showcases depth maps from various MVS methods for indoor and outdoor scenes, highlighting weakly-textured regions with lighting variations and shadows. Stereo matching based solely on photometric consistency proves challenging in these scenarios, leading to unreliable depth estimations.
\par
COLMAP exhibits poor performance in weakly-textured regions where depth maps contain numerous errors. Those errors will not be integrated to create a point cloud, resulting in a conspicuously sparse point cloud in the reconstruction. ACMP uses triangulation to generate planar priors, effectively estimating the depth of small, weakly-textured planes. However, limitations in triangulation segmentation may lead to inaccurate plane hypotheses, especially in complex scenes with occlusion or intersections of multiple planes. Although ACMMP integrating ACMP's strengths and a multi-scale image pyramid structure to enhance reconstruction performance, it still faces challenges with the impact of light and shadows.
\par
In contrast, our approach exhibits notable advantages in depth estimation for weakly-textured regions in the ETH3D datasets. Our method yields more precise depth estimations in weakly-textured regions, effectively addressing the challenge of indistinct matching. Leveraging the global information aggregation function, which incorporates comprehensive information from weakly-textured regions, our method adeptly reconstructs these challenging areas, even in the presence of lighting and shadow effects. As illustrated in \autoref{fig9}, our approach mitigates the impact of shooting angles and lighting on the depth estimation of floors and walls, substantially reducing depth map noise and enhancing overall smoothness.

\begin{table*}
\begin{center}
\caption{Quantitative comparisons of completeness, accuracy and ${F_1}$ score based on the high-resolution training and test datasets provided by the ETH3D benchmark dataset. $\tau$ is the tolerance threshold for evaluation. Completeness, accuracy and ${F_1}$ scores of other methods are taken from the ETH3D evaluation site. The best results are indicated by the bold number. }
\begin{adjustbox}{width=0.8\textwidth}
\begin{tabular}{ccccccccccc}
\toprule
& \multirow{2}*{\textbf{Methods}} & \multicolumn{3}{c}{\boldmath{$\tau=2cm$}} & \multicolumn{3}{c}{\boldmath{$\tau=5cm$}} \\
\cmidrule(lr){3-5}\cmidrule(lr){6-8}\cmidrule(lr){9-11} & &\textbf{Completeness} & \textbf{Accuracy} & \boldmath{${F_1}$} & \textbf{Completeness} & \textbf{Accuracy} & \boldmath{${F_1}$} \\ 
\midrule
\multirow{8}*{$Training$} 
&COLMAP &55.13 &\textbf{91.85} &67.66 &69.91 &\textbf{97.09} &80.50\\
&TSAR-MVS &82.27 &85.32 &83.69 &91.29 &94.25 &92.71\\
&TAPA-MVS &71.45 &85.88 &77.69 &84.83 &93.78 &88.91\\
&HC-MVS &\textbf{84.05} &75.43 &79.34 &91.82 &89.11 &90.37 \\
&QAPM &77.50 &80.43 &78.47 &87.06 &92.57 &89.49\\
&ACMP &72.15 &90.12 &79.79 &82.23 &95.96 &88.32\\
&ACMMP &77.61 &90.63 &83.42 &88.48 &96.12 &92.03\\
&Ours &83.35 &86.67 &\textbf{84.86} &\textbf{91.99} &93.89 &\textbf{92.87} \\
\midrule
\multirow{8}*{$Test$} 
&COLMAP &62.98 &\textbf{91.97} &73.01 &75.74 &\textbf{96.75} &83.96\\
&TSAR-MVS &\textbf{87.96} &83.95 &85.71 &94.35 &93.07 &93.66\\
&TAPA-MVS &74.94 &85.71 &79.15 &85.02 &92.49 &88.16\\
&HC-MVS &86.04 &83.91 &84.74 &93.71 &92.58 &93.05 \\
&QAPM &79.95 &82.59 &80.88 &86.88 &92.88 &89.51\\
&ACMP &75.58 &90.54 &81.51 &84.00 &95.71 &89.01\\
&ACMMP &81.49 &91.91 &85.89 &90.39 &96.61 &93.24\\
&Ours &87.76 &87.70 &\textbf{87.73} &\textbf{94.68} &93.84 &\textbf{94.22} \\
\midrule
\end{tabular}
\label{tb1}
\end{adjustbox}
\end{center}
\end{table*}

\begin{table*}[H]
\centering
\caption{Ablation study based on completeness, accuracy, and ${F_1}$ score for the ETH3D dataset: without Triangulation Prior (w/o TP), without SAM-based Prior (w/o SP), without Geometric Consistency For Epipolar Constraints (w/o GCEC), without Global Information Aggregation (w/o GIA) and without two of them. $\tau$ is the tolerance threshold for evaluation. The best results are indicated by the bold number.}
\begin{tabular}{ccccccccc}
\toprule
& \multirow{2}{*}{\textbf{Methods}} & \multicolumn{4}{c}{Settings} & \multicolumn{3}{c}{$\boldmath{\tau=5cm}$} \\
\cmidrule(lr){3-6}  \cmidrule(lr){7-9}
& & \textbf{TP} & \textbf{SP} & \textbf{GCEC} & \textbf{GIA} & \textbf{Completeness} & \textbf{Accuracy} & \boldmath{$F_1$} \\ 
\midrule
& ACMP (baseline) &\checkmark  &  &  &  & 82.23 & \textbf{95.96} & 88.32 \\
& w/o TP\&GCEC &  & \checkmark &  & \checkmark & 90.23 & 92.60 & 91.30 \\
& w/o TP\&GIA &  & \checkmark & \checkmark &  & 90.18 & 92.53 & 91.25 \\
& w/o SP\&GCEC & \checkmark &  &  & \checkmark & 89.05 & 91.85 & 90.32 \\
& w/o SP\&GIA &\checkmark  &  & \checkmark &  & 88.92 & 91.65 & 90.14 \\
& w/o GIA\&GCEC & \checkmark & \checkmark &  &  & 91.63 & 92.19 & 91.09 \\
& w/o TP &  & \checkmark & \checkmark & \checkmark & 91.05 & 93.67 & 92.17 \\
& w/o SP & \checkmark &  & \checkmark & \checkmark & 90.50 & 93.17 & 91.71 \\
& w/o GCEC & \checkmark & \checkmark &  & \checkmark & 91.15 & 93.62 & 92.39 \\
& w/o GIA & \checkmark & \checkmark & \checkmark &  & 90.63 & 92.95 & 91.69 \\
& Ours & \checkmark & \checkmark & \checkmark & \checkmark & \textbf{91.99} & 93.89 & \textbf{92.87} \\
\midrule
\end{tabular}
\label{tb2}
\end{table*}

\begin{figure*}[H]
  \centering
  \subfigure[training$:\tau=2cm$]{\includegraphics[width=0.47\linewidth]{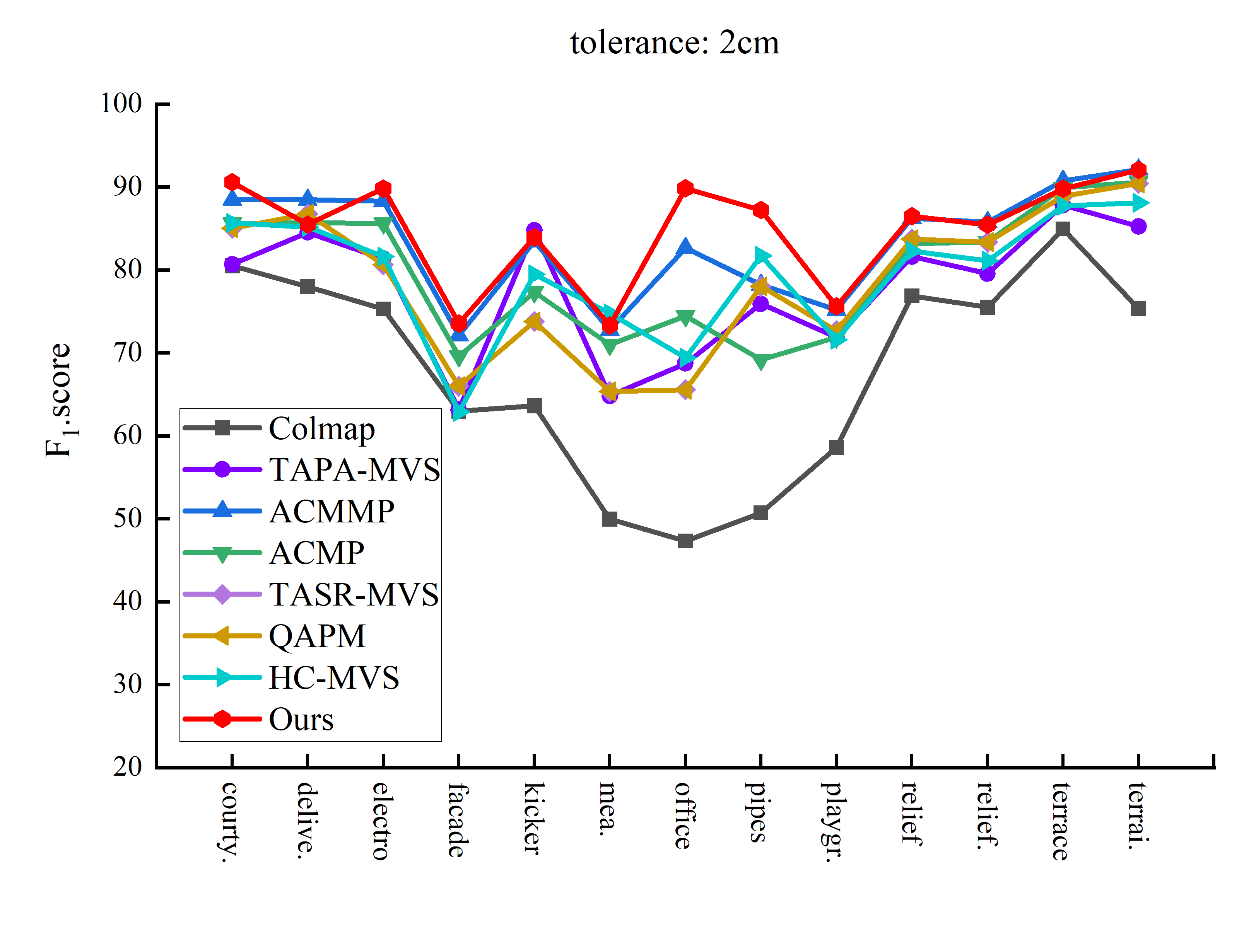}}
  \hspace{5mm}
  \subfigure[training$:\tau=5cm$]{\includegraphics[width=0.47\linewidth]{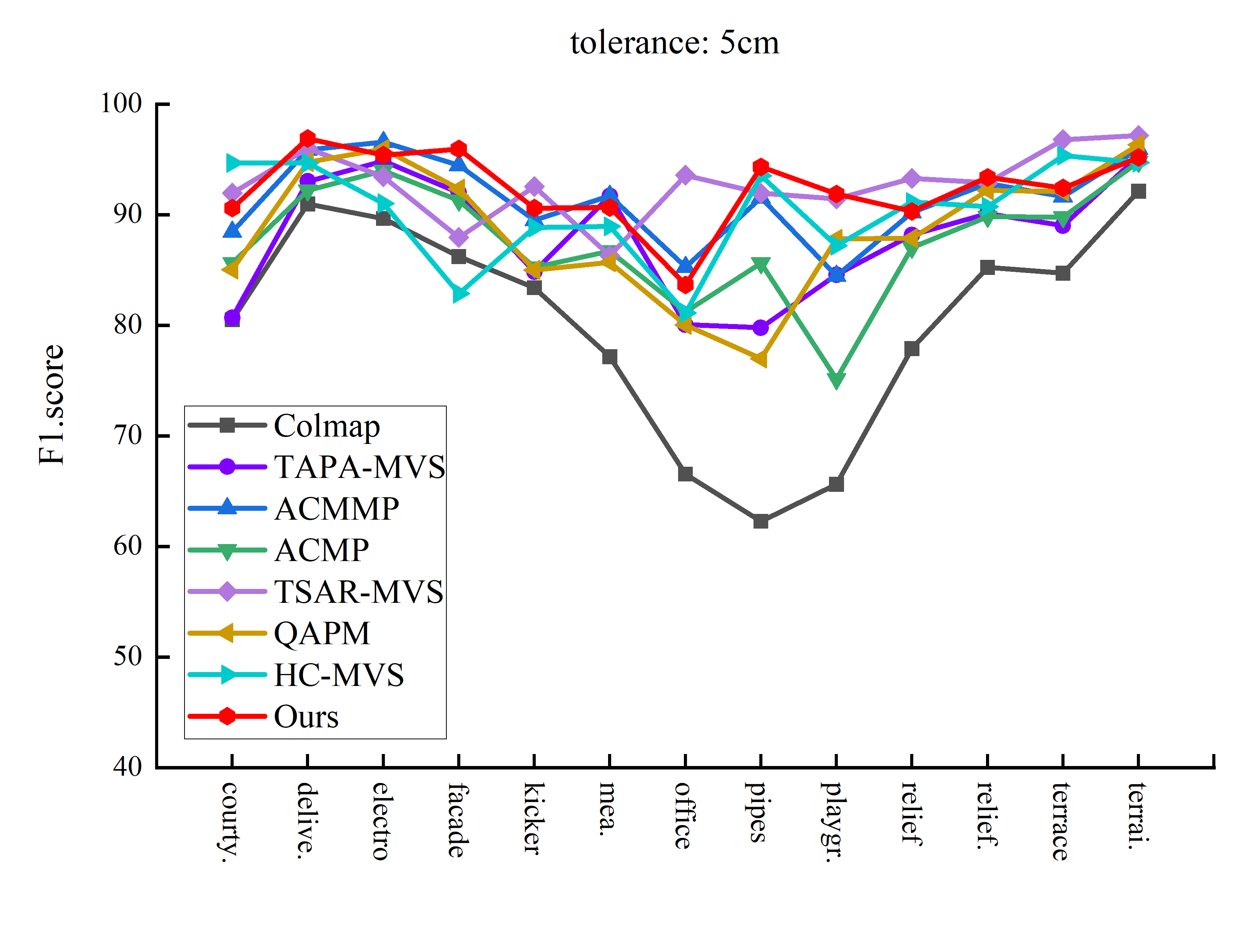}}
  \\
  \subfigure[test$:\tau=2cm$]{\includegraphics[width=0.47\linewidth]{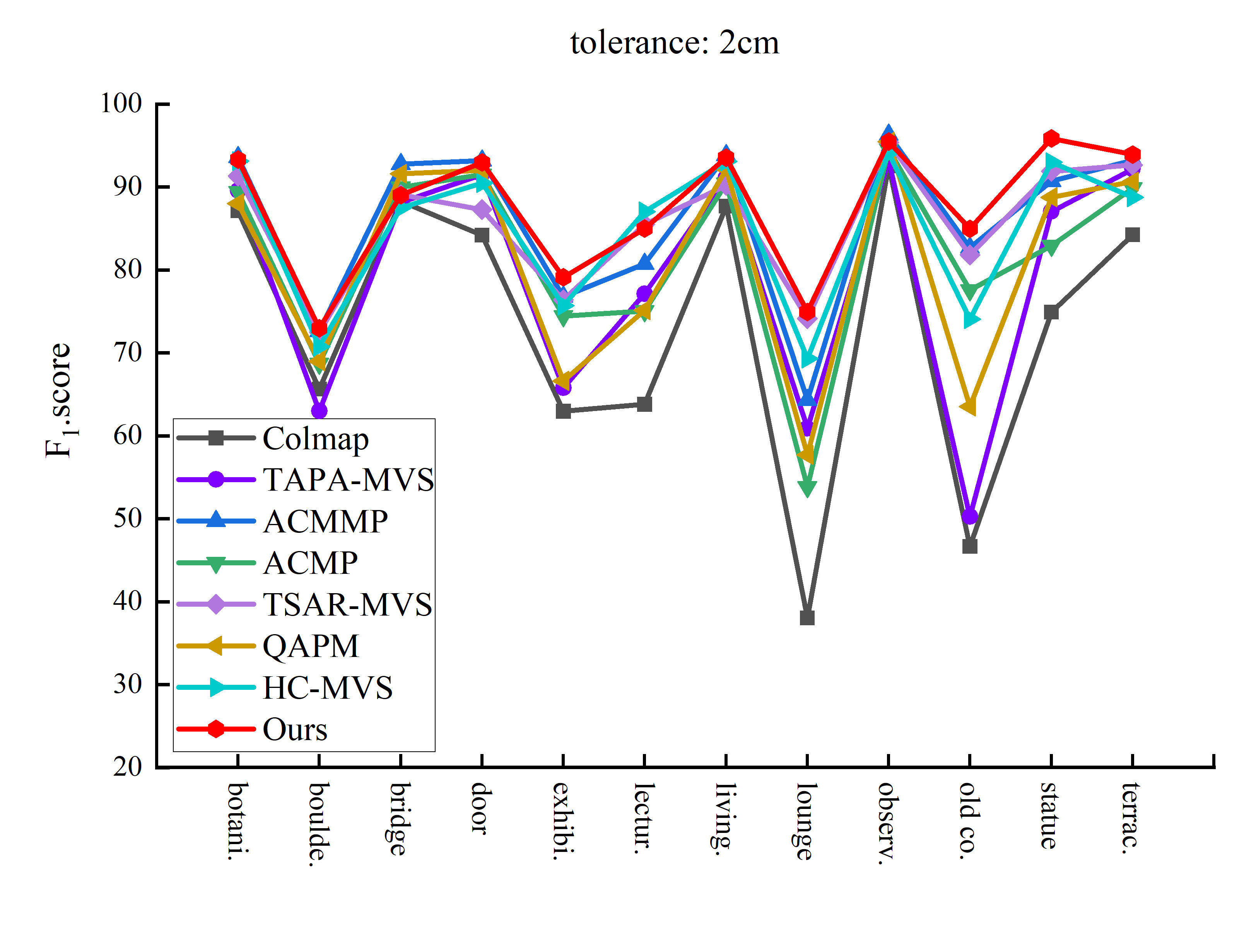}}
  \hspace{5mm}
  \subfigure[test$:\tau=5cm$]{\includegraphics[width=0.47\linewidth]{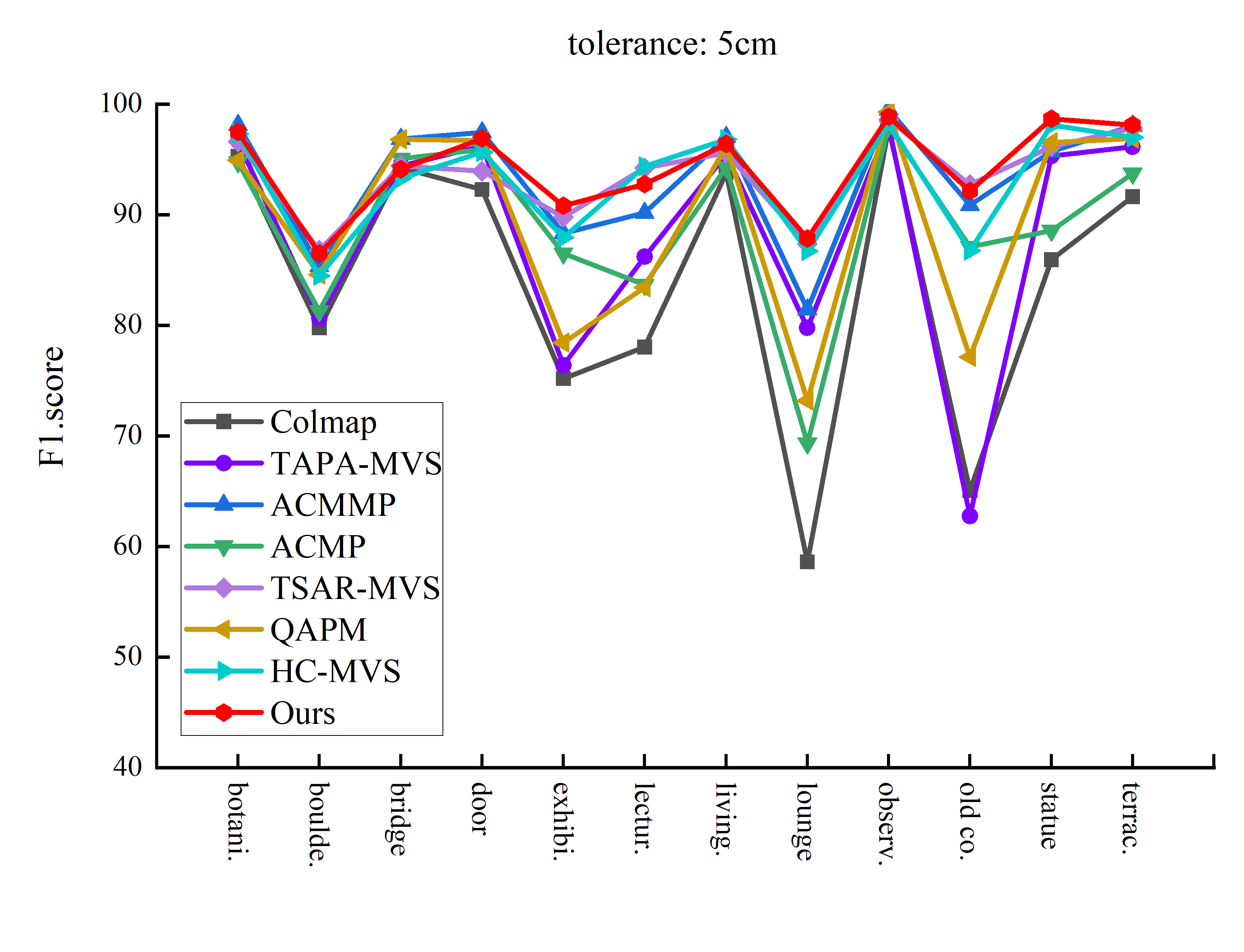}}
  \caption{Quantitative evaluation comparison results ($F_1$ score) of varying tolerances, utilizing several state-of-the-art methods on all sequences from the ETH3D benchmark dataset.}
  \label{fig12}
\end{figure*}

\begin{figure*}[h]
    \centering
    \includegraphics[width=\linewidth, height=0.7\textheight]{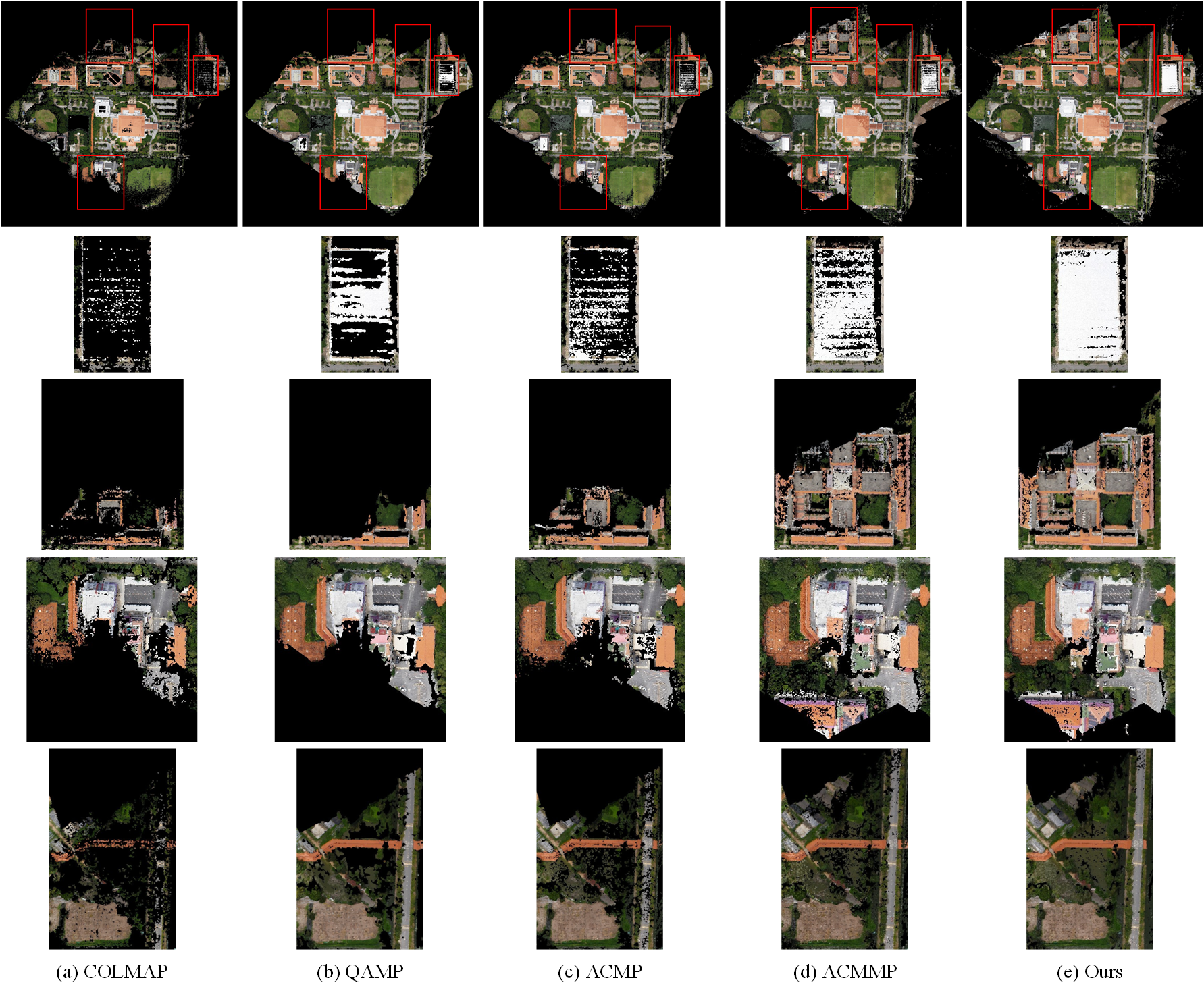}
    \caption{Qualitatively compare point clouds from various algorithms on some images from the SenseFly-university dataset: (a) COLMAP; (b) QAPM; (c) ACMP; (d) ACMMP; (e) Ours. Challenging regions are marked in red boxes and zoomed below. Black pixels in the point cloud represent areas with no data.}
    \label{fig13}
\end{figure*}

\begin{figure*}[h]
    \centering
    \includegraphics[width=\linewidth, height=0.6\textheight]{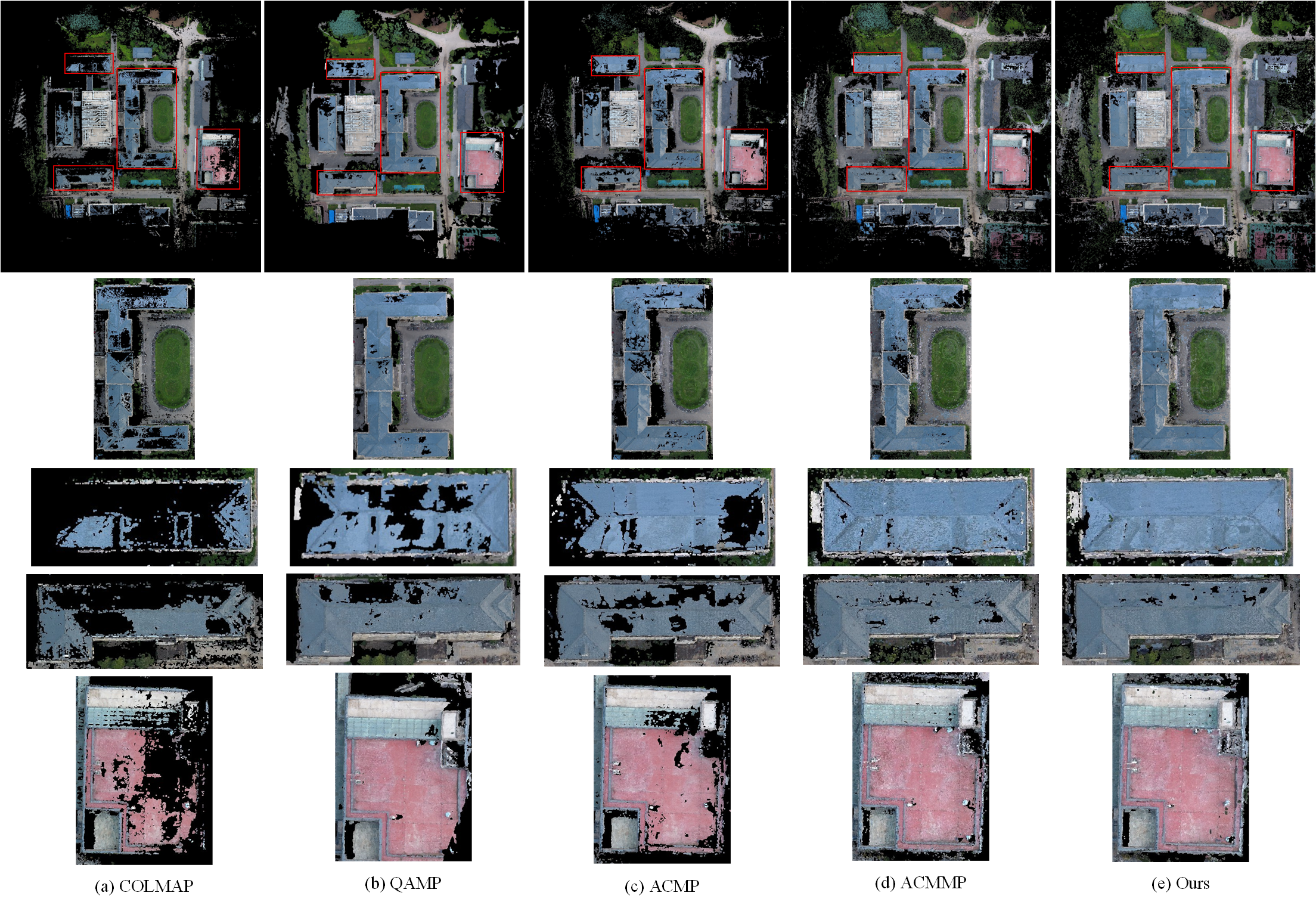}
    \caption{Qualitatively compare point clouds from various algorithms on some images from the UDD5 dataset: (a) COLMAP; (b) QAPM; (c) ACMP; (d) ACMMP; (e) Ours. Challenging regions are marked in red boxes and zoomed below. Black pixels in the point cloud represent areas with no data.}
    \label{fig14}
\end{figure*}

\begin{figure*}[h]
    \centering
    \includegraphics[width=\linewidth, height=0.3\textheight]{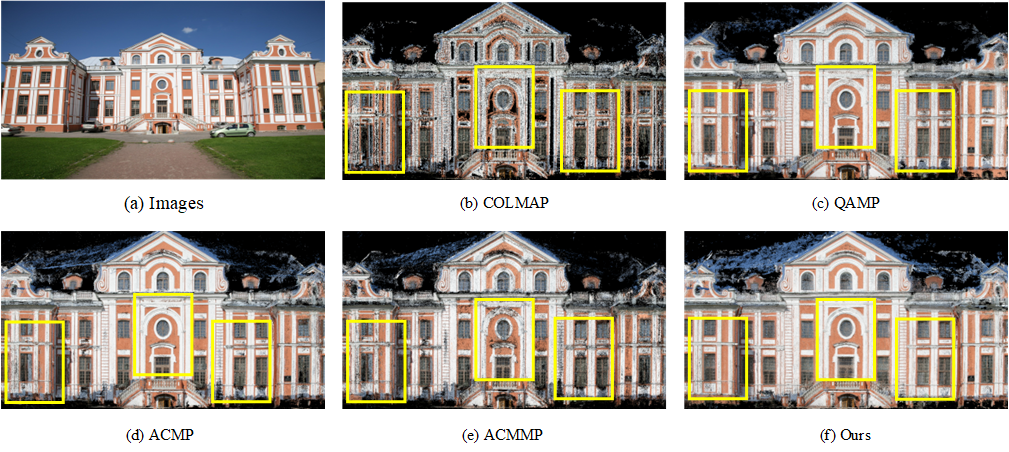}
    \caption{Qualitatively compare point clouds from various algorithms on some images from the agisoft dataset: (a) Images; (b) COLMAP; (c) QAMP; (d) ACMP; (e) ACMMP; (f) Ours. Challenging regions are marked in yellow boxes. Black pixels in the point cloud represent areas with no data.}
    \label{fig15}
\end{figure*}

\begin{figure*}[h]
    \centering
    \includegraphics[width=\linewidth, height=0.3\textheight]{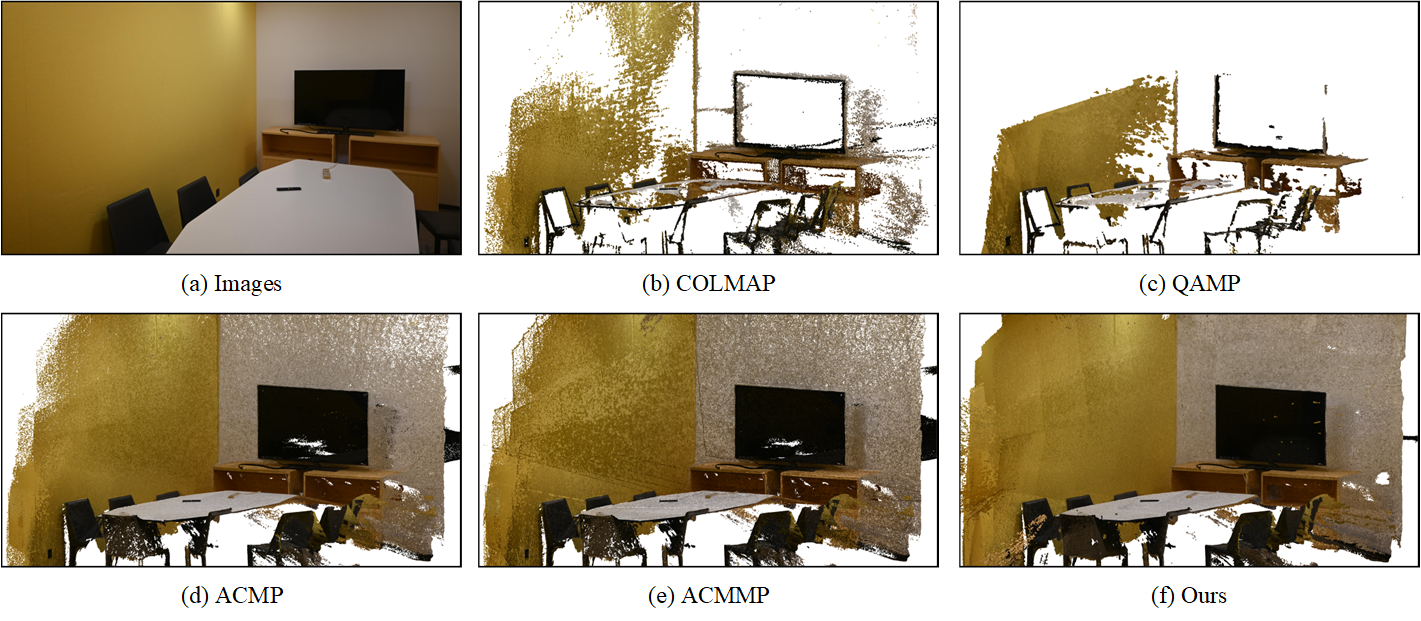}
    \caption{Qualitatively compare point clouds from various algorithms on some images from the real scenarios: (a) Images; (b) COLMAP; (c) QAMP; (d) ACMP; (e) ACMMP; (f) Ours. White pixels in the point cloud represent areas with no data.}
    \label{fig16}
\end{figure*}

\subsection{Point Cloud Evaluation}
We conducted a comprehensive analysis of the reconstructed point clouds utilize the dataset from the ETH3D benchmark, both quantitatively and qualitatively. Our method was compared with several state-of-the-art MVS methods. The quantitative results are presented in \autoref{tb1}, with $F_1$ scores for various scenes are illustrated in \autoref{fig12}. The findings highlight the consistent outperformance of our method across various depth error ranges in terms of completeness and $F_1$ score. Notably, our method's completeness score significantly surpasses that of other approaches, providing further evidence of its efficacy in reconstructing weakly-textured regions and generating point clouds with high completeness. This superiority is attributed to the reliable set of candidate priors we introduce, enabling accurate segmentation and modeling of weakly-textured regions and effectively reducing noise generation.
\par
To visually demonstrate the superiority of our method in terms of high completeness and accuracy, we conducted a qualitative analysis of the reconstructed point cloud. Qualitative comparisons of dense point clouds are shown in \autoref{fig10} and \autoref{fig11}. Our method achieved outstanding results in the reconstruction of both indoor and outdoor scenes, successfully reconstructing most of the weakly-textured regions in the scenes, including floors, walls, doors, and building surfaces.
\par
Additionally, we conducted experiments on high resolution aerial datasets, presenting point cloud reconstructions of the UDD and SenseFly datasets in \autoref{fig13} and \autoref{fig14}. Aerial dataset scene reconstruction poses a significant challenge due to low image overlap and numerous weakly-textured regions, primarily concentrated on the surfaces of man-made structures and roads.
\par
In order to further test the performance of our method for reconstructing buildings. We also conducted experiments in agisoft public dataset as well as in real scenarios, as shown in  \autoref{fig15} and \autoref{fig16}. These datasets contain indoor and outdoor scenes of buildings, with many weakly-textured regions that are affected by certain lighting, which makes it challenging to reconstruct these scenes.
\par
From the reconstruction results of the ETH3D dataset, the aerial dataset, agisoft dataset and real scenarios, it is evident that our method produces a more dense and complete point cloud compared to other methods. In \autoref{tb1}, while COLMAP achieves the highest accuracy score, its completeness is compromised, stemming from a localized accuracy emphasis, resulting in a conspicuously sparse point cloud, particularly in weakly-textured regions. The reconstruction results emphasize COLMAP's suboptimal performance, limiting its practical applicability. ACMP and QAPM, grappling with the issue of over-segmentation, can only reconstruct partial regions of weakly-textured regions. Although ACMMP mitigates this problem to some extent, the reconstructed point cloud remains incomplete. TSAR-MVS, while capable of recovering certain weakly-textured regions, introduces a notable amount of noise. While HC-MVS has greatly improved the completeness of 3D models, its accuracy is lacking. This indicates that the completion strategy of HC-MVS suffers from low completion accuracy, resulting in a significant amount of noise. As can see in point cloud qualitative from the above dataset as well as real scenarios. In contrast, our method, with a reliable prior candidate set and consideration of global information, demonstrates significant advantages in the reconstruction of weakly-textured regions. Both qualitative and quantitative analyses highlight the efficacy of our method in reconstructing more complete 3D models.

\section{Ablation}
To validate the reliability and effectiveness of each component in our proposed method, we assessed their performance on the high-resolution training set from the ETH3D benchmark. Beginning with the baseline model, we incrementally incorporated different components and evaluated their performance metrics, as depicted in \autoref{tb2}.
\par
It is evident that the absence of triangulation prior (TP) and SAM-based prior (SP) resulted in declines in completeness and $F_1$. In particular, in the absence of SP, fewer points were reconstructed in weakly-textured regions, diminishing completeness. Compared with TP, SP provides more complete prior information and solves the over segmentation problem of TP. Notably, our method shows a decrease in accuracy compared to the baseline, which is due to the introduction of prior modules. When only TP or SP are present, completeness significantly improves, but accuracy drops. Because prior information can provide more comprehensive prior plane but neglects some details of the scene's geometry. Nevertheless, our method strikes a balance between completeness and accuracy, sacrificing some accuracy to achieve substantial improvements in completeness and $F_1$ score.
\par
To address this issue and update the superior depths, both completeness and accuracy improved under the combined influence of geometric consistency for epipolar constraint (GCEC) and global information aggregation (GIA). Without geometric consistency, relying solely on photometric consistency for matching in weakly-textured regions can easily lead to local optima. Considering geometric constraints between camera viewpoints enhances matching capability in such regions by limiting the search range. Without global information aggregation cost leads to incorrect involvement of reliable prior information in depth updates, introducing many erroneous points and ultimately causing a decline in accuracy and completeness. Global aggregation allows our method to adapt better to scene complexity, enhancing robustness in handling large-scale scenes and weakly-textured regions. Compared to removing GIA and GCEC, when GIA or GCEC operates individually, completeness decreases, but accuracy increases. Under the combined influence of both, the completeness and accuracy of the point cloud improve. This is because both need to work together. Guided by global information aggregation, the aggregation cost of the current pixel considers both geometric constraints and global information, ensuring that more reliable prior information participates in the update of depth estimates.

\section{Runtime Analysis}
We execute our method on a single GPU and a single CPU. Then record the runtime for each stage and compare the total time with other state-of-the-art methods, as detailed in \autoref{tb3} and \autoref{tb4}. ACMP is the most time-efficient method. ACMMP, which employs an image pyramid structure, requires more time than ACMP. QAPM, running on a single CPU thread, incurs the highest time cost. 
\par
Our method is slower than ACMP and ACMMP primarily due to generate SAM masks and SAM-based prior generation emerge as the most time-consuming stages. The extended runtime is primarily attributed to two factors. Firstly, the computational demand of SAM for segmenting high-resolution images is considerable. Secondly, the fitting and updating of prior planes within segmented regions affect to the overall time demand. In summary, the total runtime of our method stays within a reasonable range, though further improvements for mask generation and plane fitting remain to be explored.

\begin{table}[H]
\begin{center}
\caption{Running time of different stages of the proposed method for each image of size 3200 × 2130 pixels on GPU. }
\begin{tabular}{cccccccc}
\toprule
&\textbf{Stage} & \textbf{Time$(s)$} & \textbf{Ratio$(\%)$} \\ 
\midrule
&Generate Raw Depth &22.67 &25.93\\
&Generate Triangulation Prior &1.62 &1.85\\
&Generate SAM Masks &28.83 &32.98\\
&Generate SAM-based Prior &22.28 &25.48\\
&Geometric Consistency &7.69 &8.80\\
&Update of the Hypotheis &4.34 &4.96\\
\midrule
&Total Time &87.43 &$-$\\
\midrule
\end{tabular}
\label{tb3}
\end{center}
\end{table}

\begin{table}[H]
\begin{center}
\caption{Running time of different methods for each image of size 3200 × 2130 pixels. }
\small
\begin{tabular}{ccccccc}
\toprule
&\textbf{Methods}  &QAPM  &ACMP &ACMMP &Ours\\ 
\midrule
& \textbf{Total Time$(s)$} &663.34 &\textbf{22.67}  &52.66  &87.43\\
\midrule
\end{tabular}
\label{tb4}
\end{center}
\end{table}


\section{Implications and Limitations}
Our work aims to enhance the completeness and density of 3D building reconstruction by introducing plane priors, geometric consistency for epipolar line constraints and global information aggregation. Leveraging SAM for segmenting weakly-textured regions within images empowers us to intricately guide the reconstruction of scenes, particularly in weakly-textured regions. The global information aggregation and geometric consistency for epipolar constraints integrate prior plane into the depth estimation update process to elevate the completeness and accuracy of the point cloud, culminating in a more exhaustive 3D building model.
\par
Potential areas for improvement include further optimizing the SAM and global information aggregation methods to enhance their applicability and performance across various scenarios, with a specific focus on accuracy, as current methods fit weakly-textured regions into more complete prior planes while losing some fine geometric structures, resulting in a loss of accuracy compared to our baseline. Additionally, integrating MVS in engineering applications can provide more precise and detailed 3D models for civil engineering, cultural heritage preservation, and virtual reality. Through these enhancements, our work is poised to have a greater impact on a broader range of applications, providing accuracy and comprehensive building models for urban planning, infrastructure projects, and beyond.
\section{Discussions}
We have introduced a novel approach to 3D building reconstruction that significantly advances the state of the art in handling weakly-textured regions, such as walls, floors, and roofs, which are common challenges in large-scale reconstruction tasks. Our method is the first to seamlessly integrate SAM-based planar priors with global information aggregation, ensuring high-quality depth estimation across a wide range of scenarios.
\par
A key innovation in our approach is the use of SAM and RANSAC algorithm for generating accurate planar priors. Unlike traditional methods that often suffer from over-segmentation and fragmentation, our approach leverages the robust segmentation capabilities of SAM. This allows to achieve more accurate and complete planar fittings, particularly in weakly-textured regions, which has been a long-standing challenge in the field.
\par
Moreover, our method demonstrates that by incorporating both local and global consistency through geometric constraints and global information aggregation, it is possible to significantly improve the depth estimation process. This finding challenges the prevailing assumption that local consistency alone is sufficient for high-quality depth estimation, highlighting the importance of a more holistic approach that considers the broader information of the scene.
\par
Our implementation in a probabilistic graph model for cost aggregation further demonstrates the practicality of our approach. Experiments demonstrate that our method resolves issues of incomplete reconstruction and missing point cloud in weakly-textured regions, especially in buildings with extensive weakly textured areas. In the final point cloud quality comparison, our method achieves state-of-the-art results, effectively reconstructing more complete point clouds.
\section{Conclusion}
In this study, we propose a Multi-View Stereo matching method featuring global aggregation priors. We achieve precise segmentation and modeling of prior planes within large weakly-textured regions of the building scene. Simultaneously, through the integration of triangulation priors, we establish a reliable candidate set for prior information. A novel global information aggregation cost is derived employing a probabilistic graph model, effectively incorporating optimal plane prior information into the depth estimation update process. Leveraging the planarity of planes and global information, our approach showcases strong robustness. The ETH3D dataset, the aerial dataset, agisoft dataset and real scenarios experiments demonstrate that, compared to other MVS methods, our approach excels in both depth estimation and point cloud reconstruction, highlighting its effectiveness in reconstructing weakly-textured regions within high-resolution scenes and producing models with enhanced completeness. By addressing the limitations found in previous engineering applications concerning large and complex scenes, as well as the challenge of information loss during scene reconstruction, our approach offers significant advancements in the field.
\par
Therefore, our innovative solution has great potential to overcome the limitations of 3D building reconstruction in engineering applications. Our method plays a crucial role in enhancing 3D building modeling, achieving precise engineering measurements in civil engineering by generating dense point clouds with high completeness.
\par
While our current implementation already shows substantial improvements in reconstruction quality, we recognize that there is further optimization, and our method introduces a planar prior, resulting in a certain loss of accuracy.  In the future work, more advanced segmentation model fitting model can be used to fit weak texture regions, so as to provide a more accurate and complete prior model. It can also be considered to combine the advantages of advanced deep learning algorithms and traditional algorithms to improve the accuracy and efficiency of multi-view stereo matching algorithm.
\section{Acknowledgment}
This work was supported by the Key-Area Research and Development Program of Guangdong Province, China (No.2022B0701180001) and the Science and Technology Planning Project of Guangzhou, China (No.2023B01J0007).

\printcredits

\bibliographystyle{elsarticle-harv}

\bibliography{cas-dc.bib}

\begin{thebibliography}{39}
\expandafter\ifx\csname natexlab\endcsname\relax\def\natexlab#1{#1}\fi
\providecommand{\url}[1]{\texttt{#1}}
\providecommand{\href}[2]{#2}
\providecommand{\path}[1]{#1}
\providecommand{\DOIprefix}{doi:}
\providecommand{\ArXivprefix}{arXiv:}
\providecommand{\URLprefix}{URL: }
\providecommand{\Pubmedprefix}{pmid:}
\providecommand{\doi}[1]{\href{http://dx.doi.org/#1}{\path{#1}}}
\providecommand{\Pubmed}[1]{\href{pmid:#1}{\path{#1}}}
\providecommand{\bibinfo}[2]{#2}
\ifx\xfnm\relax \def\xfnm[#1]{\unskip,\space#1}\fi
\bibitem[{Aati and Nejim(2020)}]{r49}
\bibinfo{author}{Aati, S.}, \bibinfo{author}{Nejim, S.}, \bibinfo{year}{2020}.
\newblock \bibinfo{title}{Identification of aircraft aerodynamic derivatives based on photogrammetry and computational fluid dynamics}, in: \bibinfo{booktitle}{IOP Publishing}, pp. \bibinfo{pages}{012035--}.
\bibitem[{Barnes et~al.(2009)Barnes, Shechtman, Finkelstein and Goldman}]{r15}
\bibinfo{author}{Barnes, C.}, \bibinfo{author}{Shechtman, E.}, \bibinfo{author}{Finkelstein, A.}, \bibinfo{author}{Goldman, D.B.}, \bibinfo{year}{2009}.
\newblock \bibinfo{title}{Patchmatch: A randomized correspondence algorithm for structural image editing}.
\newblock \bibinfo{journal}{TOG} \bibinfo{volume}{28}.
\newblock \DOIprefix\doi{10.1145/1531326.1531330}.
\bibitem[{Besl and Mckay(1992)}]{r53}
\bibinfo{author}{Besl, P.J.}, \bibinfo{author}{Mckay, H.D.}, \bibinfo{year}{1992}.
\newblock \bibinfo{title}{A method for registration of 3-d shapes}.
\newblock \bibinfo{journal}{IEEE Transactions on Pattern Analysis and Machine Intelligence} \bibinfo{volume}{14}, \bibinfo{pages}{239--256}.
\bibitem[{Bleyer et~al.(2011)Bleyer, Rhemann and Rother}]{r16}
\bibinfo{author}{Bleyer, M.}, \bibinfo{author}{Rhemann, C.}, \bibinfo{author}{Rother, C.}, \bibinfo{year}{2011}.
\newblock \bibinfo{title}{Patchmatch stereo - stereo matching with slanted support windows}, in: \bibinfo{editor}{Hoey, J.}, \bibinfo{editor}{McKenna, S.}, \bibinfo{editor}{Trucco, E.} (Eds.), \bibinfo{booktitle}{BMVC}.
\newblock \DOIprefix\doi{10.5244/C.25.14}.
\bibitem[{Cernea(2020)}]{r17}
\bibinfo{author}{Cernea, D.}, \bibinfo{year}{2020}.
\newblock \bibinfo{title}{{OpenMVS}: Multi-view stereo reconstruction library}.
\newblock \URLprefix \url{https://cdcseacave.github.io/openMVS}.
\bibitem[{Chen et~al.(2023)Chen, Fan and Li}]{r48}
\bibinfo{author}{Chen, S.}, \bibinfo{author}{Fan, G.}, \bibinfo{author}{Li, J.}, \bibinfo{year}{2023}.
\newblock \bibinfo{title}{Improving completeness and accuracy of 3d point clouds by using deep learning for applications of digital twins to civil structures}.
\newblock \bibinfo{journal}{Advanced Engineering Informatics} \bibinfo{volume}{58}, \bibinfo{pages}{102196}.
\newblock \URLprefix \url{https://www.sciencedirect.com/science/article/pii/S1474034623003245}, \DOIprefix\doi{https://doi.org/10.1016/j.aei.2023.102196}.
\bibitem[{Chen et~al.(2018)Chen, Wang, Lu, Chen and Wang}]{r32}
\bibinfo{author}{Chen, Y.}, \bibinfo{author}{Wang, Y.}, \bibinfo{author}{Lu, P.}, \bibinfo{author}{Chen, Y.}, \bibinfo{author}{Wang, G.}, \bibinfo{year}{2018}.
\newblock \bibinfo{title}{Large-scale structure from motion with semantic constraints of aerial images}, in: \bibinfo{editor}{Lai, J.}, \bibinfo{editor}{Liu, C.}, \bibinfo{editor}{Chen, X.}, \bibinfo{editor}{Zhou, J.}, \bibinfo{editor}{Tan, T.}, \bibinfo{editor}{Zheng, N.}, \bibinfo{editor}{Zha, H.} (Eds.), \bibinfo{booktitle}{PRCV}, pp. \bibinfo{pages}{347--359}.
\newblock \DOIprefix\doi{10.1007/978-3-030-03398-9\_30}.
\bibitem[{Cremers and Kolev(2011)}]{r12}
\bibinfo{author}{Cremers, D.}, \bibinfo{author}{Kolev, K.}, \bibinfo{year}{2011}.
\newblock \bibinfo{title}{Multiview stereo and silhouette consistency via convex functionals over convex domains}.
\newblock \bibinfo{journal}{IEEE TPAMI} \bibinfo{volume}{33}, \bibinfo{pages}{1161--1174}.
\newblock \DOIprefix\doi{10.1109/TPAMI.2010.174}.
\bibitem[{Fathi et~al.(2015)Fathi, Dai and Lourakis}]{r43}
\bibinfo{author}{Fathi, H.}, \bibinfo{author}{Dai, F.}, \bibinfo{author}{Lourakis, M.}, \bibinfo{year}{2015}.
\newblock \bibinfo{title}{Automated as-built 3d reconstruction of civil infrastructure using computer vision: Achievements, opportunities, and challenges}.
\newblock \bibinfo{journal}{Advanced Engineering Informatics} \bibinfo{volume}{29}, \bibinfo{pages}{149--161}.
\newblock \URLprefix \url{https://www.sciencedirect.com/science/article/pii/S1474034615000245}, \DOIprefix\doi{https://doi.org/10.1016/j.aei.2015.01.012}. \bibinfo{note}{infrastructure Computer Vision}.
\bibitem[{Furukawa and Ponce(2010)}]{r13}
\bibinfo{author}{Furukawa, Y.}, \bibinfo{author}{Ponce, J.}, \bibinfo{year}{2010}.
\newblock \bibinfo{title}{Accurate, dense, and robust multiview stereopsis}.
\newblock \bibinfo{journal}{IEEE TPAMI} \bibinfo{volume}{32}, \bibinfo{pages}{1362--1376}.
\newblock \DOIprefix\doi{10.1109/TPAMI.2009.161}.
\bibitem[{Goesele et~al.(2007)Goesele, Snavely, Curless, Hoppe and Seitz}]{r14}
\bibinfo{author}{Goesele, M.}, \bibinfo{author}{Snavely, N.}, \bibinfo{author}{Curless, B.}, \bibinfo{author}{Hoppe, H.}, \bibinfo{author}{Seitz, S.M.}, \bibinfo{year}{2007}.
\newblock \bibinfo{title}{Multi-view stereo for community photo collections}, in: \bibinfo{booktitle}{ICCV}, pp. \bibinfo{pages}{1--8}.
\newblock \DOIprefix\doi{10.1109/ICCV.2007.4408933}.
\bibitem[{Hirschmuller(2005)}]{r28}
\bibinfo{author}{Hirschmuller, H.}, \bibinfo{year}{2005}.
\newblock \bibinfo{title}{Accurate and efficient stereo processing by semi-global matching and mutual information}, in: \bibinfo{booktitle}{CVPR}, pp. \bibinfo{pages}{807--814 vol. 2}.
\newblock \DOIprefix\doi{10.1109/CVPR.2005.56}.
\bibitem[{Kirillov et~al.(2023)Kirillov, Mintun, Ravi, Mao, Rolland, Gustafson, Xiao, Whitehead, Berg, Lo, Doll{\'a}r and Girshick}]{r25}
\bibinfo{author}{Kirillov, A.}, \bibinfo{author}{Mintun, E.}, \bibinfo{author}{Ravi, N.}, \bibinfo{author}{Mao, H.}, \bibinfo{author}{Rolland, C.}, \bibinfo{author}{Gustafson, L.}, \bibinfo{author}{Xiao, T.}, \bibinfo{author}{Whitehead, S.}, \bibinfo{author}{Berg, A.C.}, \bibinfo{author}{Lo, W.Y.}, \bibinfo{author}{Doll{\'a}r, P.}, \bibinfo{author}{Girshick, R.}, \bibinfo{year}{2023}.
\newblock \bibinfo{title}{Segment anything}.
\newblock \bibinfo{journal}{arXiv:2304.02643} .
\bibitem[{Kuhn et~al.(2019)Kuhn, Lin and Erdler}]{r54}
\bibinfo{author}{Kuhn, A.}, \bibinfo{author}{Lin, S.}, \bibinfo{author}{Erdler, O.}, \bibinfo{year}{2019}.
\newblock \bibinfo{title}{Plane completion and filtering for multi-view stereo reconstruction} \bibinfo{volume}{11824}, \bibinfo{pages}{18--32}.
\newblock \DOIprefix\doi{10.1007/978-3-030-33676-9\_2}. \bibinfo{note}{41st DAGM German Conference on Pattern Recognition (DAGM GCPR), Dortmund, GERMANY, SEP 10-13, 2019}.
\bibitem[{Liao et~al.(2024)Liao, Zhang, Huang, Fu, Huang, Cao, Xu, Xiong and Cai}]{r51}
\bibinfo{author}{Liao, Y.}, \bibinfo{author}{Zhang, X.}, \bibinfo{author}{Huang, N.}, \bibinfo{author}{Fu, C.}, \bibinfo{author}{Huang, Z.}, \bibinfo{author}{Cao, Q.}, \bibinfo{author}{Xu, Z.}, \bibinfo{author}{Xiong, X.}, \bibinfo{author}{Cai, S.}, \bibinfo{year}{2024}.
\newblock \bibinfo{title}{High completeness multi-view stereo for dense reconstruction of large-scale urban scenes}.
\newblock \bibinfo{journal}{ISPRS Journal of Photogrammetry and Remote Sensing} \bibinfo{volume}{209}, \bibinfo{pages}{173--196}.
\newblock \URLprefix \url{https://www.sciencedirect.com/science/article/pii/S0924271624000273}, \DOIprefix\doi{https://doi.org/10.1016/j.isprsjprs.2024.01.018}.
\bibitem[{Lv et~al.(2021)Lv, Tu, Tang, Liu and Shen}]{r40}
\bibinfo{author}{Lv, M.}, \bibinfo{author}{Tu, D.}, \bibinfo{author}{Tang, X.}, \bibinfo{author}{Liu, Y.}, \bibinfo{author}{Shen, S.}, \bibinfo{year}{2021}.
\newblock \bibinfo{title}{Semantically guided multi-view stereo for dense 3d road mapping}, in: \bibinfo{booktitle}{2021 IEEE International Conference on Robotics and Automation (ICRA)}, pp. \bibinfo{pages}{11189--11195}.
\newblock \DOIprefix\doi{10.1109/ICRA48506.2021.9561077}.
\bibitem[{Ma and Liu(2018)}]{r41}
\bibinfo{author}{Ma, Z.}, \bibinfo{author}{Liu, S.}, \bibinfo{year}{2018}.
\newblock \bibinfo{title}{A review of 3d reconstruction techniques in civil engineering and their applications}.
\newblock \bibinfo{journal}{Advanced Engineering Informatics} \bibinfo{volume}{37}, \bibinfo{pages}{163--174}.
\newblock \URLprefix \url{https://www.sciencedirect.com/science/article/pii/S1474034617304275}, \DOIprefix\doi{https://doi.org/10.1016/j.aei.2018.05.005}.
\bibitem[{Mahami et~al.(2019)Mahami, Nasirzadeh, Hosseininaveh~Ahmadabadian and Nahavandi}]{r56}
\bibinfo{author}{Mahami, H.}, \bibinfo{author}{Nasirzadeh, F.}, \bibinfo{author}{Hosseininaveh~Ahmadabadian, A.}, \bibinfo{author}{Nahavandi, S.}, \bibinfo{year}{2019}.
\newblock \bibinfo{title}{Automated progress controlling and monitoring using daily site images and building information modelling}.
\newblock \bibinfo{journal}{Buildings} \bibinfo{volume}{9}.
\newblock \URLprefix \url{https://www.mdpi.com/2075-5309/9/3/70}, \DOIprefix\doi{10.3390/buildings9030070}.
\bibitem[{Romanoni and Matteucci(2019)}]{r29}
\bibinfo{author}{Romanoni, A.}, \bibinfo{author}{Matteucci, M.}, \bibinfo{year}{2019}.
\newblock \bibinfo{title}{{TAPA-MVS: Textureless-Aware PatchMatch Multi-View Stereo}}, in: \bibinfo{booktitle}{ICCV}, pp. \bibinfo{pages}{10412--10421}.
\newblock \DOIprefix\doi{10.1109/ICCV.2019.01051}.
\bibitem[{Schnabel et~al.(2007)Schnabel, Wahl and Klein}]{r27}
\bibinfo{author}{Schnabel, R.}, \bibinfo{author}{Wahl, R.}, \bibinfo{author}{Klein, R.}, \bibinfo{year}{2007}.
\newblock \bibinfo{title}{Efficient ransac for point-cloud shape detection}.
\newblock \bibinfo{journal}{COMPUT GRAPH FORUM} \bibinfo{volume}{26}, \bibinfo{pages}{214--226}.
\newblock \DOIprefix\doi{10.1111/j.1467-8659.2007.01016.x}.
\bibitem[{Schonberger et~al.(2016)Schonberger, Zheng, Frahm and Pollefeys}]{r19}
\bibinfo{author}{Schonberger, J.L.}, \bibinfo{author}{Zheng, E.}, \bibinfo{author}{Frahm, J.M.}, \bibinfo{author}{Pollefeys, M.}, \bibinfo{year}{2016}.
\newblock \bibinfo{title}{Pixelwise view selection for unstructured multi-view stereo}, in: \bibinfo{editor}{Leibe, B.}, \bibinfo{editor}{Matas, J.}, \bibinfo{editor}{Sebe, N.}, \bibinfo{editor}{Welling, M.} (Eds.), \bibinfo{booktitle}{ECCV}, pp. \bibinfo{pages}{501--518}.
\newblock \DOIprefix\doi{10.1007/978-3-319-46487-9\_31}.
\bibitem[{Sch{\"o}nberger et~al.(2016)Sch{\"o}nberger, Zheng, Frahm and Pollefeys}]{r34}
\bibinfo{author}{Sch{\"o}nberger, J.L.}, \bibinfo{author}{Zheng, E.}, \bibinfo{author}{Frahm, J.M.}, \bibinfo{author}{Pollefeys, M.}, \bibinfo{year}{2016}.
\newblock \bibinfo{title}{Pixelwise view selection for unstructured multi-view stereo}, in: \bibinfo{booktitle}{ECCV}, \bibinfo{organization}{Springer}. pp. \bibinfo{pages}{501--518}.
\bibitem[{Schöps et~al.(2017)Schöps, Schönberger, Galliani, Sattler, Schindler, Pollefeys and Geiger}]{r30}
\bibinfo{author}{Schöps, T.}, \bibinfo{author}{Schönberger, J.L.}, \bibinfo{author}{Galliani, S.}, \bibinfo{author}{Sattler, T.}, \bibinfo{author}{Schindler, K.}, \bibinfo{author}{Pollefeys, M.}, \bibinfo{author}{Geiger, A.}, \bibinfo{year}{2017}.
\newblock \bibinfo{title}{A multi-view stereo benchmark with high-resolution images and multi-camera videos}, in: \bibinfo{booktitle}{CVPR}, pp. \bibinfo{pages}{2538--2547}.
\newblock \DOIprefix\doi{10.1109/CVPR.2017.272}.
\bibitem[{Seitz et~al.(2006)Seitz, Curless, Diebel, Scharstein and Szeliski}]{r35}
\bibinfo{author}{Seitz, S.}, \bibinfo{author}{Curless, B.}, \bibinfo{author}{Diebel, J.}, \bibinfo{author}{Scharstein, D.}, \bibinfo{author}{Szeliski, R.}, \bibinfo{year}{2006}.
\newblock \bibinfo{title}{A comparison and evaluation of multi-view stereo reconstruction algorithms}, in: \bibinfo{booktitle}{2006 IEEE Computer Society Conference on Computer Vision and Pattern Recognition (CVPR'06)}, pp. \bibinfo{pages}{519--528}.
\newblock \DOIprefix\doi{10.1109/CVPR.2006.19}.
\bibitem[{Seitz and Dyer(1997)}]{r37}
\bibinfo{author}{Seitz, S.}, \bibinfo{author}{Dyer, C.}, \bibinfo{year}{1997}.
\newblock \bibinfo{title}{Photorealistic scene reconstruction by voxel coloring}, in: \bibinfo{booktitle}{Proceedings of IEEE Computer Society Conference on Computer Vision and Pattern Recognition}, pp. \bibinfo{pages}{1067--1073}.
\newblock \DOIprefix\doi{10.1109/CVPR.1997.609462}.
\bibitem[{Sinha et~al.(2007)Sinha, Mordohai and Pollefeys}]{r36}
\bibinfo{author}{Sinha, S.N.}, \bibinfo{author}{Mordohai, P.}, \bibinfo{author}{Pollefeys, M.}, \bibinfo{year}{2007}.
\newblock \bibinfo{title}{Multi-view stereo via graph cuts on the dual of an adaptive tetrahedral mesh}, in: \bibinfo{booktitle}{2007 IEEE 11th International Conference on Computer Vision}, pp. \bibinfo{pages}{1--8}.
\newblock \DOIprefix\doi{10.1109/ICCV.2007.4408997}.
\bibitem[{Stathopoulou et~al.(2023)Stathopoulou, Battisti, Cernea, Georgopoulos and Remondino}]{r31}
\bibinfo{author}{Stathopoulou, E.K.}, \bibinfo{author}{Battisti, R.}, \bibinfo{author}{Cernea, D.}, \bibinfo{author}{Georgopoulos, A.}, \bibinfo{author}{Remondino, F.}, \bibinfo{year}{2023}.
\newblock \bibinfo{title}{Multiple view stereo with quadtree-guided priors}.
\newblock \bibinfo{journal}{ISPRS-J. Photogramm} \bibinfo{volume}{196}, \bibinfo{pages}{197--209}.
\bibitem[{Stathopoulou and Remondino(2019)}]{r21}
\bibinfo{author}{Stathopoulou, E.K.}, \bibinfo{author}{Remondino, F.}, \bibinfo{year}{2019}.
\newblock \bibinfo{title}{Multi-view stereo with semantic priors}, in: \bibinfo{editor}{GonzalezAguilera, D.}, \bibinfo{editor}{Remondino, F.}, \bibinfo{editor}{Toschi, I.}, \bibinfo{editor}{RodriguezGonzalvez, P.}, \bibinfo{editor}{Stathopoulou, E.} (Eds.), \bibinfo{booktitle}{CIPA}, pp. \bibinfo{pages}{1135--1140}.
\newblock \DOIprefix\doi{10.5194/isprs-archives-XLII-2-W15-1135-2019}.
\bibitem[{Vogiatzis et~al.(2007)Vogiatzis, Esteban, Torr and Cipolla}]{r11}
\bibinfo{author}{Vogiatzis, G.}, \bibinfo{author}{Esteban, C.H.}, \bibinfo{author}{Torr, P.H.S.}, \bibinfo{author}{Cipolla, R.}, \bibinfo{year}{2007}.
\newblock \bibinfo{title}{Multiview stereo via volumetric graph-cuts and occlusion robust photo-consistency}.
\newblock \bibinfo{journal}{IEEE TPMI} \bibinfo{volume}{29}, \bibinfo{pages}{2241--2246}.
\newblock \DOIprefix\doi{10.1109/TPAMI.2007.70712}.
\bibitem[{Wang et~al.(2020)Wang, Galliani, Vogel, Speciale and Pollefeys}]{r52}
\bibinfo{author}{Wang, F.}, \bibinfo{author}{Galliani, S.}, \bibinfo{author}{Vogel, C.}, \bibinfo{author}{Speciale, P.}, \bibinfo{author}{Pollefeys, M.}, \bibinfo{year}{2020}.
\newblock \bibinfo{title}{Patchmatchnet: Learned multi-view patchmatch stereo} .
\bibitem[{Wu et~al.(2017)Wu, Liu, Ye, Xu and Zheng}]{r38}
\bibinfo{author}{Wu, P.}, \bibinfo{author}{Liu, Y.}, \bibinfo{author}{Ye, M.}, \bibinfo{author}{Xu, Z.}, \bibinfo{author}{Zheng, Y.}, \bibinfo{year}{2017}.
\newblock \bibinfo{title}{Geometry guided multi-scale depth map fusion via graph optimization}.
\newblock \bibinfo{journal}{IEEE Transactions on Image Processing} \bibinfo{volume}{26}, \bibinfo{pages}{1315--1329}.
\newblock \DOIprefix\doi{10.1109/TIP.2017.2651383}.
\bibitem[{Xu et~al.(2022)Xu, Kong, Tao and Pollefeys}]{r20}
\bibinfo{author}{Xu, Q.}, \bibinfo{author}{Kong, W.}, \bibinfo{author}{Tao, W.}, \bibinfo{author}{Pollefeys, M.}, \bibinfo{year}{2022}.
\newblock \bibinfo{title}{Multi-scale geometric consistency guided and planar prior assisted multi-view stereo}.
\newblock \bibinfo{journal}{TPAMI} .
\bibitem[{Xu and Tao(2019)}]{r5}
\bibinfo{author}{Xu, Q.}, \bibinfo{author}{Tao, W.}, \bibinfo{year}{2019}.
\newblock \bibinfo{title}{Multi-scale geometric consistency guided multi-view stereo}.
\newblock \bibinfo{journal}{CVPR} .
\bibitem[{Xu and Tao(2020)}]{r8}
\bibinfo{author}{Xu, Q.}, \bibinfo{author}{Tao, W.}, \bibinfo{year}{2020}.
\newblock \bibinfo{title}{Planar prior assisted patchmatch multi-view stereo}.
\newblock \bibinfo{journal}{AAAI} .
\bibitem[{Xu et~al.(2020)Xu, Liu, Shi, Wang and Zheng}]{r39}
\bibinfo{author}{Xu, Z.}, \bibinfo{author}{Liu, Y.}, \bibinfo{author}{Shi, X.}, \bibinfo{author}{Wang, Y.}, \bibinfo{author}{Zheng, Y.}, \bibinfo{year}{2020}.
\newblock \bibinfo{title}{{MARMVS: Matching Ambiguity Reduced Multiple View Stereo for Efficient Large Scale Scene Reconstruction}}, in: \bibinfo{booktitle}{2020 IEEE/CVF Conference on Computer Vision and Pattern Recognition (CVPR)}, pp. \bibinfo{pages}{5980--5989}.
\newblock \DOIprefix\doi{10.1109/CVPR42600.2020.00602}.
\bibitem[{Yang and Shi(2014)}]{r55}
\bibinfo{author}{Yang, J.}, \bibinfo{author}{Shi, Z.}, \bibinfo{year}{2014}.
\newblock \bibinfo{title}{Image-based 3d semantic modeling of building facade}, in: \bibinfo{editor}{Chmielewski, L.J.}, \bibinfo{editor}{Kozera, R.}, \bibinfo{editor}{Shin, B.S.}, \bibinfo{editor}{Wojciechowski, K.} (Eds.), \bibinfo{booktitle}{Computer Vision and Graphics}, \bibinfo{publisher}{Springer International Publishing}, \bibinfo{address}{Cham}. pp. \bibinfo{pages}{661--671}.
\bibitem[{Yuan et~al.(2023)Yuan, Cao, Jiang, Wang and Li}]{r33}
\bibinfo{author}{Yuan, Z.}, \bibinfo{author}{Cao, J.}, \bibinfo{author}{Jiang, H.}, \bibinfo{author}{Wang, Z.}, \bibinfo{author}{Li, Z.}, \bibinfo{year}{2023}.
\newblock \bibinfo{title}{{TSAR-MVS: Textureless-aware segmentation and correlative refinement guided multi-view stereo}}.
\newblock \bibinfo{journal}{arXiv preprint arXiv:2308.09990} .
\bibitem[{Zheng et~al.(2014)Zheng, Dunn, Jojic and Frahm}]{r18}
\bibinfo{author}{Zheng, E.}, \bibinfo{author}{Dunn, E.}, \bibinfo{author}{Jojic, V.}, \bibinfo{author}{Frahm, J.M.}, \bibinfo{year}{2014}.
\newblock \bibinfo{title}{Patchmatch based joint view selection and depthmap estimation}, in: \bibinfo{booktitle}{CVPR}, pp. \bibinfo{pages}{1510--1517}.
\newblock \DOIprefix\doi{10.1109/CVPR.2014.196}.
\bibitem[{Zhou et~al.(2024)Zhou, Zhu, Zhang and Du}]{r50}
\bibinfo{author}{Zhou, X.}, \bibinfo{author}{Zhu, Q.}, \bibinfo{author}{Zhang, Q.}, \bibinfo{author}{Du, Y.}, \bibinfo{year}{2024}.
\newblock \bibinfo{title}{The full-field displacement intelligent measurement of retaining structures using uav and 3d reconstruction}.
\newblock \bibinfo{journal}{Measurement} \bibinfo{volume}{227}, \bibinfo{pages}{114311}.
\newblock \URLprefix \url{https://www.sciencedirect.com/science/article/pii/S0263224124001957}, \DOIprefix\doi{https://doi.org/10.1016/j.measurement.2024.114311}.

\end{thebibliography}



\end{document}